\documentclass[sigconf]{acmart}

\usepackage{subfig}
\usepackage{multirow}

\AtBeginDocument{%
  }

\copyrightyear{2024}
\acmYear{2024}
\setcopyright{acmlicensed}\acmConference[WWW '24]{Proceedings of the ACM Web Conference 2024}{May 13--17, 2024}{Singapore, Singapore}
\acmBooktitle{Proceedings of the ACM Web Conference 2024 (WWW '24), May 13--17, 2024, Singapore, Singapore}
\acmDOI{10.1145/3589334.3645406}
\acmISBN{979-8-4007-0171-9/24/05}

\begin{document}

\title{A Knowledge-Injected Curriculum Pretraining Framework for Question Answering}

\author{Xin Lin}
\orcid{0000-0001-6913-4654}
\affiliation{%
  \institution{School of Computer Science and Technology, University of Science and Technology of China \& State Key Laboratory of Cognitive Intelligence}
  \city{Hefei}
  \country{China}
}
\email{linx@mail.ustc.edu.cn}

\author{Tianhuang Su}
\orcid{0009-0001-1195-3195}
\affiliation{%
  \institution{Guangdong OPPO Mobile Telecommunications Corp., Ltd}
  \city{Shenzhen}
  \country{China}
}
\email{sutianhuang@oppo.com}

\author{Zhenya Huang}
\authornote{Corresponding Author.}
\orcid{0000-0003-1661-0420}
\affiliation{%
  \institution{School of Computer Science and Technology, University of Science and Technology of China \& State Key Laboratory of Cognitive Intelligence}
  \city{Hefei}
  \country{China}
}
\email{huangzhy@ustc.edu.cn}

\author{Shangzi Xue}
\orcid{0009-0004-6426-9647}
\affiliation{%
  \institution{School of Computer Science and Technology, University of Science and Technology of China \& State Key Laboratory of Cognitive Intelligence}
  \city{Hefei}
  \country{China}
}
\email{xueshangzi@mail.ustc.edu.cn}

\author{Haifeng Liu}
\orcid{0009-0000-2922-3898}
\affiliation{%
  \institution{University of Science and Technology of China}
  \city{Hefei}
  \country{China}
}
\email{bladehliu@qq.com}

\author{Enhong Chen}
\orcid{0000-0002-4835-4102}
\affiliation{%
  \institution{Anhui Province Key Laboratory of Big Data Analysis and Application, University of Science and Technology of China \& State Key Laboratory of Cognitive Intelligence}
  \city{Hefei}
  \country{China}
}
\email{cheneh@ustc.edu.cn}

\renewcommand{\shortauthors}{Xin Lin et al.}

\begin{abstract}
  Knowledge-based question answering (KBQA) is a key task in natural language processing research, and also an approach to access the web data and knowledge, which requires exploiting knowledge graphs (KGs) for reasoning. In the literature, one promising solution for KBQA is to incorporate the pretrained language model (LM) with KGs by generating KG-centered pretraining corpus, which has shown its superiority. However, these methods often depend on specific techniques and resources to work, which may not always be available and restrict its application. Moreover, existing methods focus more on improving language understanding with KGs, while neglect the more important human-like complex reasoning. To this end, in this paper, we propose a general \textbf{K}nowledge-\textbf{I}njected \textbf{C}urriculum \textbf{P}retraining framework (KICP) to achieve comprehensive KG learning and exploitation for KBQA tasks, which is composed of knowledge injection (KI), knowledge adaptation (KA) and curriculum reasoning (CR). Specifically, the KI module first injects knowledge into the LM by generating KG-centered pretraining corpus, and generalizes the process into three key steps that could work with different implementations for flexible application. Next, the KA module learns knowledge from the generated corpus with LM equipped with an adapter as well as keeps its original natural language understanding ability to reduce the negative impacts of the difference between the generated and natural corpus. Last, to enable the LM with complex reasoning, the CR module follows human reasoning patterns to construct three corpora with increasing difficulties of reasoning, and further trains the LM from easy to hard in a curriculum manner to promote model learning. We provide an implementation of the general framework, and evaluate the proposed KICP on four real-word datasets. The results demonstrate that our framework can achieve higher performances, and have good generalization ability to other QA tasks.
\end{abstract}

  



\maketitle

\section{Introduction}

\begin{figure}[t]
  \centering
  \includegraphics[width=\linewidth]{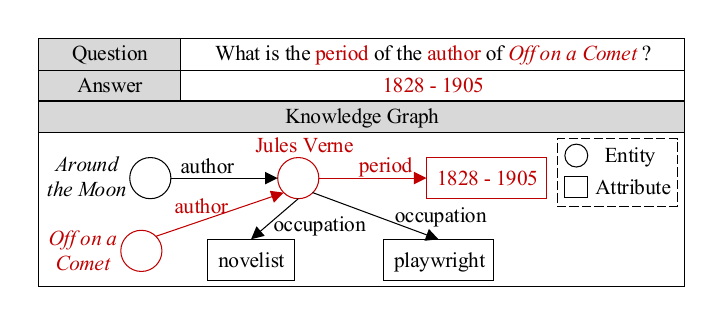}
  \caption{A toy example of KBQA, which requires complex reasoning marked in red.}
  \Description{A toy example of KBQA, which requires complex reasoning marked in red.}
  \label{fig:example}
\end{figure}

Knowledge-based question answering (KBQA) is a key task in natural language processing (NLP) and data mining research~\cite{saxena2020improving}, which could act as an approach to access and process web data and knowledge, and lead to useful applications such as smart voice assistant and search engine especially with the large language models (LLMs)~\cite{ouyang2022training}. As shown in Figure~\ref{fig:example}, KBQA aims to answer questions in natural language based on background knowledge, which is often formatted as knowledge graphs (KGs)~\cite{yasunaga2021qa, zhang2022greaselm, liu2023knowledge}. Therefore, KBQA requires abilities of both natural language understanding (NLU) and knowledge reasoning, making it a challenging task in related fields. 

In the literature, researchers have proposed many solutions for KBQA~\cite{saxena2020improving, lv2020graph, zhang2022greaselm} based on deep learning due to their remarkable results on other NLP tasks~\cite{huang2021disenqnet, lin2023learning, liu2023learning}, among which the pretrained language models (LMs) have become the most promising for its strong NLU ability~\cite{devlin2019bert, ouyang2022training, chen2022knowprompt, meng2022topic}. Unfortunately, LMs including the LLMs work not so well in knowledge application~\cite{logan2019barack, liu2020k}, which hinders its application in KBQA. Therefore, researchers have tried great efforts to enhance the LMs with KGs (inputting knowledge facts into LMs, or pretraining LMs with knowledge-based tasks~\cite{liu2020k, zhang2019ernie, peters2019knowledge, sun2019ernie, wang2021kepler, wang2021k, yu2022jaket, zhu2023knowledge, ye2022ontology}), which has greatly improved LMs in knowledge-related tasks. However, these methods often learn KGs as supplementary to additional pretraining corpus~\cite{zhang2019ernie, liu2020k}, which can not cover the whole KG and may overlook some knowledge useful in certain tasks, and thus leads to incomplete knowledge learning. Towards this point, a straightforward solution is to generate the pretraining corpus based on the KGs. Although many methods have been developed along this line~\cite{liu2022enhancing, agarwal2021knowledge, zhang2023structure, chen2020kgpt}, they usually depend on specific techniques or resources for effective corpus generation (e.g., requiring pretrained generative model to generate sentences, or generating sentences in a fixed format), which may be unavailable in practice and thus restricts its application. Therefore, in this paper we hope to design a general framework to generate KG-centered corpus for comprehensive knowledge pretraining of LMs, which is not limited to specific techniques and could work with different detailed implementations for flexible application.

However, along this line there exist several nontrivial technical challenges. First, there are many solutions to generate sentences based on given KGs for different demands (e.g., pretrained generative LMs~\cite{agarwal2021knowledge}, fixed sentence templates~\cite{liu2022enhancing}). Moreover, although most KGs store the knowledge triples with entity IDs, some high-quality KGs also contain additional attribute information, which is stored in various forms (e.g., texts, numbers and dates) and requires different processing. How to unify and generalize these various techniques and data forms remains much open. Second, the generated sentences differ from natural ones and may even seem distorted, which may mislead the LM and hurt natural language understanding ability of the LM in pretraining~\cite{agarwal2021knowledge, liu2022enhancing}. Existing methods address this problem with specific techniques in accordance with their generation methods (e.g., generating sentences more similar to natural ones with complex generative LMs~\cite{agarwal2021knowledge}, or adopting specially designed sentence templates to reduce the negative impacts~\cite{liu2022enhancing}), but how to overcome this shortcoming for an arbitrary generation method in the general framework is a nontrivial problem. Last, existing methods enhancing LMs with KGs focus more on improving language understanding with related knowledge such as K-BERT~\cite{liu2020k} and ERNIE~\cite{zhang2019ernie}, while seldom have considered the human-like complex reasoning ability. Humans can perform reasoning over multiple knowledge facts following specific patterns, which is also widely required in KBQA tasks. For example, in Figure~\ref{fig:example}, to reach the answer, the LM first needs to find that the author of \textit{Off on a Comet} is Jules Verne, and then the period of Jules Verne is 1828-1905. How to enable the LMs with such complex reasoning is a challenging problem.

To this end, in this paper, we propose a general \textbf{K}nowledge-\textbf{I}njected \textbf{C}urriculum \textbf{P}retraining framework (KICP) to achieve comprehensive KG learning and exploitation for KBQA, which is composed of knowledge injection (KI), knowledge adaptation (KA) and curriculum reasoning (CR). Specifically, the KI module converts KG triples into sentences to construct pretraining corpus for complete knowledge learning, and generalizes the process into three key steps, i.e., text characterization, sentence construction and masking, which can be implemented with different detailed techniques and various data forms for flexible application. Next, to reduce the negative impacts brought by the difference between generated and natural corpus on LM pretraining, the KA module fixes the original LM to keep its NLU ability, and learns knowledge from the generated corpus with a trainable adapter working with the LM. Last, to pretrain the LM with complex reasoning ability, the CR module follows common reasoning patterns of humans and constructs corpora requiring complex knowledge reasoning. Furthermore, the CR module arranges the complex corpora into three lessons with increasing difficulties, and trains the LM from easy to hard following the curriculum learning manner to reduce pretraining difficulty. Finally, we provide an implementation of the general framework, and conduct extensive experiments on four real-word datasets to evaluate KICP. The results demonstrate that our framework can achieve higher performances, and generalize to other QA tasks well.

\begin{figure*}[t]
  \centering
  \includegraphics[width=\linewidth]{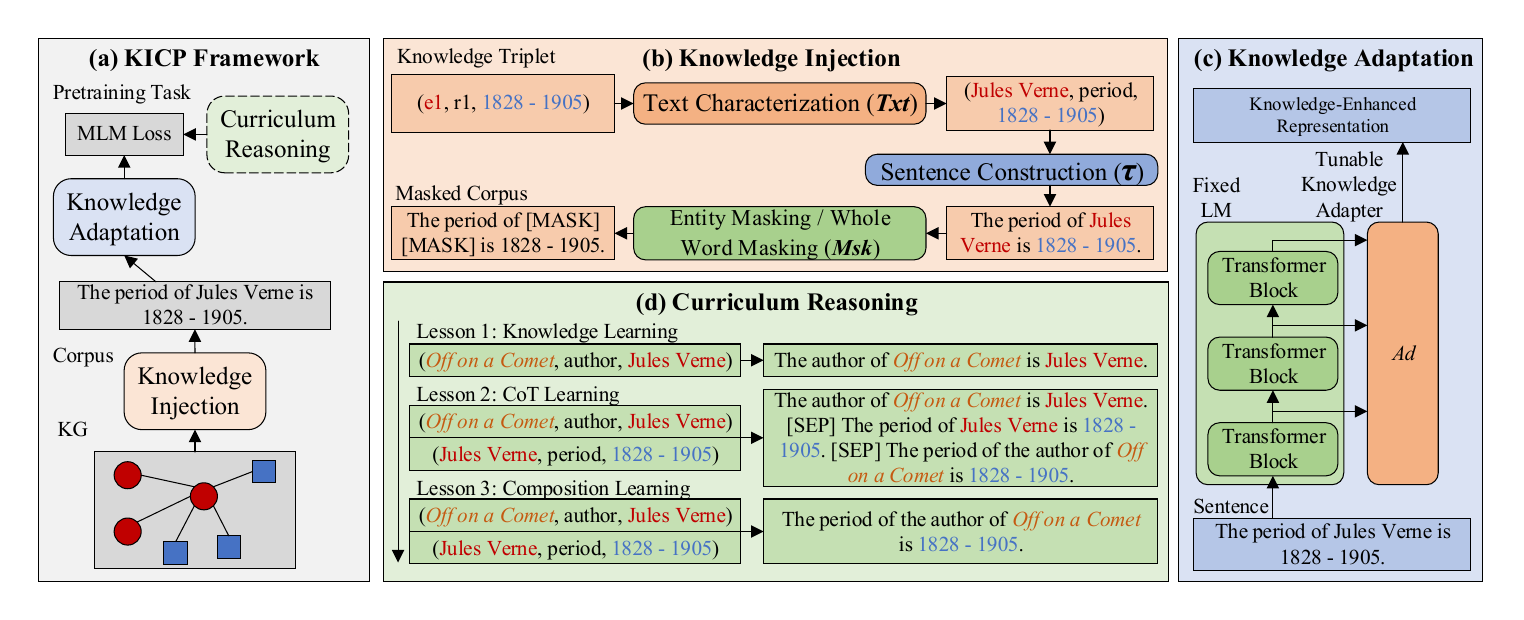}
  \caption{The architecture of the proposed KICP framework. (a) The overview of KICP. (b) The knowledge injection module (KI) converts KG triples into sentences. (c) The knowledge adaptation module (KA) works with the LM to keep NLU ability and learn knowledge. (d) The curriculum reasoning module (CR) constructs easy-to-hard reasoning-required pretraining corpora.}
  \Description{The architecture of the proposed KICP framework. (a) The overview of KICP. (b) The knowledge injection module (KI) converts KG triples into sentences. (c) The knowledge adaptation module (KA) works with the LM to keep NLU ability and learn knowledge. (d) The curriculum reasoning module (CR) constructs easy-to-hard reasoning-required pretraining corpora.}
  \label{fig:framework}
\end{figure*}

\section{Related Work}

\noindent{\textbf{Knowledge-Based Question Answering.}}
% Knowledge-based question answering (KBQA) aims to answer questions based on given knowledge bases, which are usually knowledge graphs (KGs)~\cite{saxena2020improving, lv2020graph, zhang2022greaselm, liu2023knowledge}. 
In the literature, studies on KBQA can be roughly divided into the knowledge-enhanced LM (introduced later), and the KG-based reasoning including path-based~\cite{lukovnikov2019pretrained}, embedding-based~\cite{saxena2020improving, huang2019knowledge} and graph-based methods~\cite{lv2020graph, lin2019kagnet, yasunaga2021qa, zhang2022greaselm, hu2022empowering, yasunaga2022deep}. Path-based methods map the question into entities and relations for reasoning on the KG~\cite{lukovnikov2019pretrained}.
% , which have higher interpretability but require much effort on rule design in complex questions. 
Embedding-based methods such as EmbedKGQA~\cite{saxena2020improving} represent the question and KG in the same latent space, and infer the answer with simple vector computation. 
% These methods unify simple and complex reasoning, but may have limited performance and interpretability. 
Graph-based methods~\cite{lv2020graph, lin2019kagnet, yasunaga2021qa, zhang2022greaselm, hu2022empowering, yasunaga2022deep} sample a sub-graph from the KG, and perform reasoning on the sub-graph with neural networks. Graph-based methods are widely applied in complex reasoning for the good trade-off between interpretability and performance, but the insufficient knowledge modeling within the sub-graph may lead to limited robustness. Besides, the large language models (LLMs) have become a promising method in KBQA tasks recently~\cite{achiam2023gpt,du2022glm}. Researchers have proposed several advanced techniques to improve its knowledge reasoning ability, including the chain-of-thought prompt~\cite{wei2022chain}, question decomposition~\cite{zhou2022least}, and retrieval augmented generation~\cite{yao2022react}.

\noindent{\textbf{Knowledge-Enhanced Language Model.}}
% As the pretrained language models demonstrate great performances in natural language processing~\cite{vaswani2017attention, ouyang2022training, chen2022knowprompt, meng2022topic}, their shortcomings on knowledge-based tasks are also exposed~\cite{logan2019barack, liu2020k}. Therefore, 
As the pretrained LMs have shown its weakness on knowledge-based tasks~\cite{logan2019barack, liu2020k}, researchers have tried many efforts to enhance LMs with knowledge from KGs, including the explicit methods~\cite{liu2020k, zhang2019ernie, peters2019knowledge, lu2022kelm} and implicit methods~\cite{sun2019ernie, wang2021kepler, wang2021k, xiongpretrained, liu2022enhancing, feng2023factkb}. Explicit methods feed knowledge facts or embeddings into LM as additional inputs to exploit related knowledge. For example, K-BERT~\cite{liu2020k} injected the knowledge triples into the sentences as inputs to the LM. Zhang \textit{et al.}~\cite{zhang2019ernie} developed an aggregator network to incorporate KG entity embeddings into LMs. Implicit methods design special pretraining tasks to learn knowledge from KGs and corpus with LM. Sun \textit{et al.}~\cite{sun2019ernie} introduced an entity masking strategy for pretraining, and Wang \textit{et al.}~\cite{wang2021kepler} trained LM as knowledge embedding model with entity descriptions. To better exploit KG triples, Liu \textit{et al.}~\cite{liu2022enhancing} generated multilingual synthetic pretraining corpus with KG triples and Agarwal \textit{et al.}~\cite{agarwal2021knowledge} employed the generative LM to synthesize more natural corpus. In summary, explicit methods exploit the knowledge more directly but require additional knowledge annotations as inputs, while implicit methods can be easily applied in downstream tasks, but require heavy pretraining.

% \noindent{\textbf{Curriculum Learning.}}
% Curriculum learning is an effective continual optimization strategy first proposed by Bengio~\cite{bengio2009curriculum}, which imitates human learning habits starting by easy lessons and then more difficult ones, and demonstrates that training model on datasets from easy to hard could benefit learning, accelerate convergence and promote the training outcome. Curriculum learning has shown great superiority in improving the generalization and convergence of models, and has been widely applied in various fields~\cite{zhao2022jiuzhang, li2020competence, maoneuro}. For example, Zhao \textit{et al.}~\cite{zhao2022jiuzhang} designed pretraining tasks with different difficulties and applied curriculum learning to train a LM for mathematics understanding, and Li \textit{et al.}~\cite{li2020competence} trained the visual question solver on a sequence of instance sets with increasing complexity following the curriculum manner.

Our work differs from previous methods as follows. First, existing methods converting the KG into corpus are often limited to specific techniques and resources, while our method is a general framework which can work with different detailed implementations for different circumstances. Second, existing methods focus more on improving language understanding with related knowledge, while our method further enables the LM with complex reasoning ability with specially designed pretraining task.

\section{KICP: Knowledge-Injected Curriculum Pretraining}

% In this section, we first formally introduce the KBQA task, and then present the proposed KICP framework.

\subsection{Problem Definition}

Knowledge-based question answering (KBQA) is composed of the knowledge graph $\mathcal{KG}$ and the question-answer pair $(Q, Y)$. We suppose that the KG contains knowledge triples about the relation between two entities and the attribute of each entity, where the attribute values are in diverse forms that can be converted into texts (texts are defined as $V^+$ on vocabulary $V$). Therefore, the KG can be defined as $\mathcal{KG}=(\mathbb{E}, \mathbb{R}, \sum)$, where $\mathbb{E}$ is the entity set, $\mathbb{R}$ is the relation and attribute set, and $\sum$ means the knowledge triples. 
Each triple $(h, r, t)\in\sum$ ($h, t\in\mathbb{E}$, $r\in\mathbb{R}$) means that the entity $h$ and $t$ have the relation $r$ (e.g., ``Jules Verne'' is the ``author'' of ``\textit{Off on a Comet}'' in Figure~\ref{fig:example}), and $(h, r, t)\in\sum$ ($h\in\mathbb{E}$, $r\in\mathbb{R}$, $t\in V^+$) means the attribute $r$ of entity $h$ is $t$, where $t$ is the attribute value in text (e.g., the ``period'' of ``Jules Verne'' is ``1828-1905'') 
Besides, each entity $e\in\mathbb{E}$ is assigned with several names $N_e=\{{n_e}_1, {n_e}_2, \dots, {n_e}_k\}$ (each name ${n_e}_i\in V^+$). $\mathbb{R}$ is assigned with names similarly. 
In the question-answer pair $(Q, Y)$, $Q=\{q_1, q_2, \dots, q_n\}\in V^+$ ($q_i\in V$) is the question in natural language, and $Y$ is the answer to $Q$ inferred under $\mathcal{KG}$, whose form depends on the task (e.g., KBQA often selects an entity or attribute value from $\mathcal{KG}$, and generative QA generates formal language from certain vocabulary such as natural text or mathematical expression~\cite{liu2022cognitive, liu2023guiding}). 

Given the knowledge graph $\mathcal{KG}$ and question-answer pair $(Q, Y)$, the goal of KBQA is to train a model $\mathit{M:}(\mathcal{KG}, Q)\mathit{\to}Y$ to predict the answer $Y$ of question $Q$ under $\mathcal{KG}$. In this paper, we first pretrain a language model $\mathcal{LM}$ with $\mathcal{KG}$, and then use it in $\mathit{M}$ to predict the answer $Y$ to $Q$. We expect that $\mathcal{LM}$ could learn knowledge from $\mathcal{KG}$ comprehensively and well handle complex reasoning.

\subsection{Method}

We propose a general \textbf{K}nowledge-\textbf{I}njected \textbf{C}urriculum \textbf{P}retraining framework (KICP) to pretrain $\mathcal{LM}$ for comprehensive knowledge learning and complex reasoning, which is not limited to specific techniques and could easily work with different implementations for flexible applications. As shown in Figure~\ref{fig:framework}~(a), KICP is composed of three key components, i.e., \emph{knowledge injection} (KI), \emph{knowledge adaptation} (KA) and \emph{curriculum reasoning} (CR). Specifically, KI injects knowledge from the KG into the LM completely by converting the KG triples to sentences to construct the pretraining corpus, and generalize the various generation techniques into three key steps. To reduce the negative impacts brought by the gap between generated and natural corpus, KA fixes the original LM to keep its NLU ability, and equips the framework with a trainable knowledge adapter to learn knowledge from the generated corpus. To pretrain the LM with complex reasoning ability, CR follows common patterns of human reasoning and constructs several reasoning-required corpora with different difficulties, and trains the LM from easy to hard in a curriculum manner to promote model learning. 

\subsubsection{Knowledge Injection}\label{sec:knowledge_injection}

To overcome the insufficient knowledge learning brought by using the KG as supplementary to external corpus, we directly convert the KG triples into sentences as pretraining corpus to inject knowledge into the LM. Moreover, there exist several effective sentence generation techniques for different requirements in the literature~\cite{liu2022enhancing,agarwal2021knowledge}, and the KGs contain multiple forms of data that requires different processing (e.g., IDs, texts, numbers and dates). Therefore, to generalize these detailed techniques to a general framework that is not limited to specific techniques for flexible application in various circumstances, as shown in Figure~\ref{fig:framework}~(b), we abstract the sentence generation process into three key steps, i.e., text characterization, sentence construction and masking.

\noindent{\textbf{Text Characterization.}} Given one triple $k=(h, r, t)\in\sum$ sampled from $\mathcal{KG}$, KI first characterizes all fields of the triple as texts ($\mathit{Txt}$), which serve as the backbone elements of the sentence to generate. For the entities and relations stored in IDs, We map the meaningless ID (e.g., e1) to a meaningful name (Jules Verne), which is dynamically sampled from the associated name set in each iteration to increase corpus diversity. More sampling strategies can also be applied here for other demands~\cite{liu2022enhancing}. For the various forms of attribute values (e.g, numbers, dates and texts), we use their textual descriptions as they can always be expressed with texts despite the original forms. In this way, we can unify the diverse processing of the entities, relations and attribute values.

\noindent{\textbf{Sentence Construction.}} After getting the textual elements, KI applies a sentence construction strategy $\tau$ to assemble these elements into a complete sentence, including reordering and transforming the elements and adding auxiliary words. The strategy $\tau$ can be implemented with different existing techniques, such as sentence templates, grammar-based rules, and the generative LMs~\cite{liu2022enhancing, agarwal2021knowledge}.

\noindent{\textbf{Masking.}} The last step is to mask the generated sentence for masked language model (MLM) pretraining. To force knowledge learning and match the differences between entities and attribute values, we prefer paying more weights to the knowledge elements in the sentence (those converted from the triple), and applying different masking strategies $\mathit{Msk}$ to entities and attribute values. For example, we apply the entity masking~\cite{sun2019ernie} on entities which masks the whole entity name to force learning relation knowledge instead of memorizing the entity name, and whole word masking (WWM)~\cite{cui2021pre} on attribute values since the values may contain too much information (e.g., biography) and are too hard to recover if all masked. WWM also works similarly to entity masking on short values (e.g., numbers) by masking as a whole word. More masking techniques can be used here as $\mathit{Msk}$.

Overall, the sentence generation process is formulated as follows:
\begin{equation}
  \mathit{KI}(k)=\mathit{Msk}(\tau(\mathit{Txt}(h), \mathit{Txt}(r), \mathit{Txt}(t))),\ \ k=(h, r, t)\in\sum.
\end{equation}
The knowledge-injected corpus is composed of the sentences $\mathit{KI}(k)$, which are dynamically generated from triples sampled from the KG in pretraining. In this way, KI converts the whole KG into the corpus, and thus implicitly stores all information from the KG in the corpus such as the structural infromation. Compared with existing methods rewriting KG as corpus, KI does not depend on specific techniques or resources, and thus could work with different implementations for various application demands.

\subsubsection{Knowledge Adaptation}\label{sec:knowledge_adaptation}

Obviously the corpus generated by KI differs from natural ones as the sentences may not strictly follow the grammar (especially for some simple $\tau$), and the diversity of the corpus is limited. Pretraining the LM on the corpus may hurt NLU ability and work badly on natural texts. Furthermore, as the sentence generation technique in the proposed general framework is arbitrary, we can not use methods associated with specific generation techniques to address the problem as existing studies~\cite{liu2022enhancing, agarwal2021knowledge}. Therefore, in knowledge adaptation (KA), we turn to keeping the NLU ability of LM during knowledge pretraining. 

As demonstrated by Figure~\ref{fig:framework}~(c), following the adapter paradigm in LM tuning~\cite{wang2021k, ding2022delta}, we fix the LM parameters and add a trainable knowledge adapter module $\mathit{Ad}$ above the original LM $\mathit{LM}$. $\mathit{Ad}$ uses the semantic outputs of $\mathit{LM}$ as inputs, and outputs the knowledge-enhanced representations. Moreover, to deeply improve the fusion of the semantics and knowledge, the semantic outputs of all layers in the LM are used. The computation of KA is formulated as follows:
\begin{equation}
  \mathit{KA}(x)=\mathit{Ad}(\mathit{LM}(x)),
\end{equation}
where $x$ is the input sentence. $\mathit{Ad}$ can be implemented with any neural networks, which is expected to have a proper size to contain enough space for knowledge learning and avoid greatly increasing computation complexity as well.

In pretraining, the parameters of $\mathit{Ad}$ is trained to learn knowledge from the constructed corpus, while the original LM is fixed. As the original LM is not affected by $\mathit{Ad}$, the NLU ability is retained as much as possible to reduce the negative impacts of the gap between generated and natural corpus.

\subsubsection{Curriculum Reasoning}

With KI and KA, KICP can effectively inject the KG into LM, but still lacks complex reasoning ability over multiple knowledge facts as required in real-world KBQA tasks. To enable the LM with such ability, the curriculum reasoning module (CR) pretrains LM on corpora requiring complex reasoning as shown in Figure~\ref{fig:framework}~(d).

It is hard to collect enough reasoning-required corpus for all KGs, so we also build the corpus based on the KG. Humans often perform complex reasoning following specific patterns (e.g., multi-top reasoning), which put restrictions on the participating triples (e.g., the chain-like triples). Therefore, we build the corpus following these patterns (e.g., ``The period of the author of \textit{Off on a Comet} is 1828-1905''). We first sample several triples $\{k_1,\dots,k_n\}$ matching the restrictions from KG, such as the chain-like triples \{(\textit{Off on a Comet}, author, Jules Verne), (Jules Verne, period, 1828-1905)\} for multi-hop reasoning, and then convert them into a complex composition with a pipeline $\mathit{Comp}$ similar to KI as follows:
\begin{equation}
  \begin{split}
    \mathit{Comp}(k_1, \dots, k_n) = &\mathit{Msk'}(\tau'(\mathit{Txt}(h_1), \mathit{Txt}(r_1), \mathit{Txt}(t_1), \\ &\dots, \mathit{Txt}(t_n))),\ \ k_i=(h_i, r_i, t_i)\in\sum,
  \end{split}
\end{equation}
where $\tau'$ and $\mathit{Msk'}$ are sentence construction and masking in $\mathit{Comp}$. In this way, the complex corpus matches human reasoning, and explicitly exploits the structural information from the KG as well. Much more reasoning patterns can be supported by the CR module.

The complex composition often discards some information to infer from knowledge, so it is hard to pretrain LM directly (e.g., in previous example ``Jules Verne'' is discarded, which makes it hard to understand without related knowledge). Therefore, as shown in Figure~\ref{fig:framework}~(d), we split the pretraining into three lessons with generated corpora from easy to hard following curriculum learning~\cite{zhao2022jiuzhang} to promote model learning.

\noindent{\textbf{Lesson 1: Knowledge Learning.}}\label{sec:knowledge_learning}
We start by pretraining LM on single triples from the KG. We build this corpus with KI based on one triple $k$ for each sentence, and pretrain the LM (i.e., KA) on the MLM task to memorize the knowledge facts as follows:
\begin{equation}
  \min_{\theta_\mathit{Ad},\ \theta_\mathit{MLM}} L_1(k) = \mathit{MLM}(\mathit{KA}(\mathit{KI}(k))),
\end{equation}
where $\theta_\mathit{Ad}$ and $\theta_\mathit{MLM}$ means trainable parameters for knowledge adapter $\mathit{Ad}$ in $\mathit{KA}$ and MLM head.

\noindent{\textbf{Lesson 2: CoT Learning.}}\label{sec:cot_learning}
Having learned basic knowledge facts from KG, next we teach the LM how to conduct complex reasoning with related knowledge facts. Inspired by chain-of-thought (CoT)~\cite{weichain, lu2022dynamic}, we assemble each sentence with complex composition by $\mathit{Comp}$ for certain reasoning pattern and all related knowledge by $\mathit{KI}$ as reasoning steps base on triples $\{k_1,\dots,k_n\}$. To avoid information leakage, we mask the same element (e.g., entity) in both the final composition and reasoning steps, and pretrain the LM on the MLM task as follows:
\begin{equation}
  \begin{split}
  \min_{\theta_\mathit{Ad},\ \theta_\mathit{MLM}} L_2(k_1,\dots,k_n)=&\mathit{MLM}(\mathit{KA}([\mathit{KI}(k_1),\dots,\\& \mathit{KI}(k_n),\mathit{Comp}(k_1,\dots,k_n)])),
  \end{split}
\end{equation}
where $[,]$ means text concatenation, and $\{k_1, \dots, k_n\}$ matches the reasoning pattern for $\mathit{Comp}$.

\noindent{\textbf{Lesson 3: Composition Learning.}}\label{sec:composition_learning}
In the hardest lesson, we pretrain the LM to reason with memorized knowledge as real-world QA tasks, where we only provide the final compositions without related reasoning steps. Therefore, We construct the corpus with the complex compositions by $\mathit{Comp}$, and pretrain the LM on the MLM task as follows:
\begin{equation}
  \min_{\theta_\mathit{Ad},\ \theta_\mathit{MLM}} L_3(k_1,\dots,k_n) = \mathit{MLM}(\mathit{KA}(\mathit{Comp}(k_1,\dots,k_n))).
\end{equation}
The corpora are dynamically generated with randomly sampled triples in pretraining. 
We demonstrate some samples of corpora in three lessons in Appendix~\ref{app:corpus}. 
Through the three pretraining lessons, we explicitly enable the LM with human-like complex reasoning ability required in KBQA tasks, and reduce the pretraining difficulty with the curriculum learning.

\subsubsection{QA Fine-Tuning}\label{sec:fine_tuning}

After pretrained on the KG, the LM can be easily applied in different downstream QA tasks without additional annotations or external knowledge inputs. Specifically, the LM (i.e., $\mathit{KA}$) reads the question $Q$ as input, and outputs the knowledge-enhanced vector, which is fed to a task-dependent prediction head $\mathit{Pred}$ to generate the answer $Y$. The whole system ($\mathit{LM}$ and $\mathit{Ad}$ in $\mathit{KA}$ and $\mathit{Pred}$) can be fine-tuned on different QA tasks subject to the task-dependent objective function $\mathcal{L}$ as follows:
\begin{equation}
  \min_{\theta_\mathit{LM},\ \theta_\mathit{Ad},\ \theta_\mathit{Pred}} L_\mathit{QA}(Q, Y) = \mathcal{L}(\mathit{Pred}(\mathit{KA}(Q)), Y),
\end{equation}
where $\theta_\mathit{LM}$, $\theta_\mathit{Ad}$ and $\theta_\mathit{Pred}$ are parameters of these modules.

\begin{table*}
  \caption{Overall Results of All Methods on Four Datasets}
  \label{tab:overall_results}
  \centering
  \begin{tabular}{c|cc|cc|c|c}
    \toprule
    Dataset & \multicolumn{2}{c|}{CN-QA} & \multicolumn{2}{c|}{ComplexWebQuestions} & FreebaseQA & Math23K \\
    Metric & F1 & EM & F1 & EM & ACC & ACC \\
    \midrule
    GPT4 &	0.459 &	0.358 &	0.802 &	0.721 &	\textbf{0.918} & / \\
    ChatGLM2-6B &	0.389 &	0.274 &	0.494 &	0.432 &	0.610 & / \\
    \midrule
    EmbedKGQA & 0.417 &	0.303 &	0.760 &	0.730 &	0.707 & / \\
    \midrule
    BERT & 0.607 & 0.458 & 0.856 & 0.763 & 0.896 & 0.801 \\
    RoBERTa & 0.610 & 0.456 & 0.863 & 0.779 & 0.892 & 0.803 \\
    \midrule
    ERNIE & 0.614 & 0.459 & 0.861 & 0.772 & 0.901 & 0.796 \\
    K-BERT & 0.620 & 0.462 & 0.866 & 0.774 & 0.896 & 0.799 \\
    KEPLER & 0.628 & 0.467 & 0.868 & 0.785 & 0.906 & / \\
    K-Adapter & 0.612 & 0.462 & 0.866 & 0.802 & 0.905 & / \\
    \midrule
    KICP-KA & 0.633 & 0.469 & 0.871 & 0.809 & 0.903 & 0.797 \\
    KICP-ATT & 0.629 & 0.466 & / & / & / & / \\
    KICP & \textbf{0.639*} & \textbf{0.480*} & \textbf{0.880*} & \textbf{0.819*} & 0.911* & \textbf{0.809*} \\
    \bottomrule
  \end{tabular}
\end{table*}

\subsection{Implementation}\label{sec:implementation}

In this section, we provide an implementation of the general KICP framework. In KI, we implement text characterization and masking as mentioned in section~\ref{sec:knowledge_injection}, and realize $\tau$ by simply concatenating all fields, which works well on our datasets. 

In KA, we implement the knowledge adapter $\mathit{Ad}$ as BERT with the same number of layers and halved vector dimension. In each layer of $\mathit{Ad}$, the input (semantic vector from corresponding layer of $\mathit{LM}$) is first projected with a linear model to the latent space of hidden vector from last layer, and then added with the hidden vector to feed to the BERT layer. The final vectors of $\mathit{Ad}$ and $\mathit{LM}$ are merged with a linear layer as the output.
%The architecture of KA is available in Appendix~\ref{app:implementation}.

In CR, we implement $\mathit{Comp}$ with two widely-used reasoning patterns, i.e., multi-hop reasoning and multi-object reasoning. \textbf{Multi-hop reasoning} (e.g., the period of the author of \textit{Off on a Comet} is 1828-1905) first infers an intermediate entity from the topic entity in the question (the author of \textit{Off on a Comet} is Jules Verne), and then use it to infer the next intermediate entity until reaching the answer (the period of Jules Verne is 1828-1905). Therefore, the knowledge triples form a chain-like structure, where the tail entity of one triple is the head of the next one (e.g., Jules Verne). Given these triples, $\mathit{Comp}$ discards all intermediate entities and concatenates other fields sequentially. \textbf{Multi-object reasoning} (e.g., the occupation of Jules Verne is novelist and playwright) infers several results from one topic entity, thus the knowledge triples share the same head entity and relation (Jules Verne and occupation). Given the triples, $\mathit{Comp}$ discards the heads and relations expect the first one, and concatenates all tails with the first head and relation. 
Besides, our framework could also easily generalize to other reasoning patterns such as the comparative reasoning in the similar way by defining the sampling restrictions and $\mathit{Comp}$ methods for triples. 
For each sentence we sample 2 to 3 triples matching the patterns.

\section{Experiments}

% In this section, we conduct experiments on four QA datasets to evaluate the proposed KICP framework. 

\subsection{Experimental Setup}

\subsubsection{Datasets}
We use three KBQA datasets to evaluate KICP on knowledge-based reasoning, i.e., CN-QA (with CN-KG as KG), ComplexWebQuestions~\cite{talmor2018web} and FreebaseQA~\cite{jiang2019freebaseqa} (both with Wikidata~\cite{wang2021kepler}), and a generative dataset Math23K~\cite{wang2017deep} (with HowNet~\cite{qi2019openhownet}) for generalization to other knowledge-related QA.
The introduction and statistics of the datasets are available in Appendix~\ref{app:dataset}.

KBQA answers questions with entities or attribute values from KG. To reduce computation complexity without losing much difficulty, we sample 10 hard candidate answers with the same type of the truth for prediction on KBQA. We also sample a sub-graph from the whole KG for each dataset to accelerate pretraining.

\subsubsection{Baseline Methods}
We compare KICP with original LMs \textbf{BERT}~\cite{devlin2019bert} and \textbf{RoBERTa}~\cite{liu2019roberta}, and knowledge-enhanced LMs \textbf{ERNIE}~\cite{zhang2019ernie}, \textbf{K-BERT}~\cite{liu2020k}, \textbf{KEPLER}~\cite{wang2021kepler} and \textbf{K-Adapter}~\cite{wang2021k}. We also include the embedding-based \textbf{EmbedKGQA}~\cite{saxena2020improving} and two LLMs \textbf{GPT4}~\cite{achiam2023gpt} and \textbf{ChatGLM2}~\cite{du2022glm} as baselines for KBQA datasets. 
We provide a brief introduction to baselines in Appendix~\ref{app:baseline}.

\subsubsection{Training Details.}
We implement KICP with Pytorch based on pretrained BERT by huggingface.~\footnote{https://huggingface.co/transformers} We use the ``bert-base-chinese'' version as $\mathit{LM}$ on Chinese datasets CN-QA and Math23K, and ``bert-base-uncased'' on English datasets ComplexWebQuestions and FreebaseQA for all methods. The number of BERT layers of $\mathit{Ad}$ for KA is 12 (equal to $\mathit{LM}$), the dimension is 384 for hidden vector (half of $\mathit{LM}$) and 768 for output vector(equal to $\mathit{LM}$). 
% The parameters of $\mathit{Ad}$ are initialized with kaiming initialization. The implementations of KI and CR are available in section~\ref{sec:implementation}.

We pretrain the model for 3 epochs with AdamW~\cite{loshchilovdecoupled}. The batch size is set to 32, and the learning rate is 0.0005, which warms up over the first 10\% steps, and then linearly decays. The masking probability for MLM is set to 0.15 in lesson 1 and 3, and 0.3 in lesson 2 as the corpus contains more repeated information.

% In downstream QA tasks, for CN-QA, ComplexWebQuestions and FreebaseQA, we concatenate the question and each candidate answer as input to LM and implement the classifier $\mathit{Pred}$ with MLP. CN-QA and ComplexWebQuestions are viewed as multi-label classification with more than one answers for each question and fine-tuned with binary cross entropy loss, and FreebaseQA is fine-tuned with cross entropy loss as single-label classification. For Math23K, we input the question into LM as encoder, and adopt GTS~\cite{xie2019goal}, an effective MWP solver, as decoder, which is fine-tuned with cross entropy loss. The QA dataset is much smaller than the KG, thus we fine-tune for 30 epochs on CN-QA, ComplexWebQuestions and FreebaseQA, and 80 epochs on more difficult Math23K.

We run all experiments on a Linux server with two 2.20 GHz Intel Xeon E5-2650 CPUs and a Tesla K80 GPU.~\footnote{Our codes are available at https://github.com/l-xin/KICP.}

\subsection{Experimental Results}

\subsubsection{Overall Results}

In this section, we compare KICP with all baselines. We use the F1 score (F1) and exact match score (EM) as metrics for multi-label datasets CN-QA and ComplexWebQuestions, and accuracy (ACC) for single-label dataset FreebaseQA. Math23K is evaluated with answer accuracy (ACC), i.e., the predicted expression is viewed correct if the computed answer equals the truth.

The results on four datasets are reported in Table~\ref{tab:overall_results}.~\footnote{We do not evaluate KEPLER and K-Adapter on Math23K, as pretraining the two methods requires entity descriptions, which are unavailable on HowNet.} We statistically test the improvement of KICP over baselines (except GPT4) with paired t-test, and find the improvement to be significant with $p < 0.05$ (marked *). We can get the following observations. First, KICP outperforms all baselines, which clearly demonstrates its effectiveness on knowledge learning and exploitation for QA tasks. Second, KICP performs better than K-Adapter with similar model but different pretraining task, showing the significant influence of pretraining task. Third, LLMs do not perform better than the fine-tuned methods on KBQA. GPT4 achieves comparable performance on the widely studied ComplexWebQuestions and FreebaseQA, but falls far behind on CN-QA, and the smaller ChatGLM2 performs even worse.
% Fourth, the embedding-based EmbedKGQA does not work well due to its limited representation ability.
Fourth, knowledge-enhanced methods outperform original LMs in most cases, proving that knowledge is a key element in QA reasoning. Last, knowledge injection does not bring much improvement and even negative effect on Math23K. The reason may be that Math23K requires NLU much more than knowledge.
% , which may be affected by knowledge injection and thus hurts reasoning.

\subsubsection{Ablation Study}

% Besides the widely studied entity relation knowledge stored in IDs, KICP further incorporates the attribute knowledge in diverse forms. Moreover, KICP designs the knowledge adapter module to reduce the negative impacts of the generated corpus. Therefore, in this section, we conduct ablation experiments to study the effectiveness of the two components 
In this section, we conduct ablation experiments to study the effectiveness of the attribute knowledge and knowledge adaptation 
(We will investigate the curriculum reasoning in detail in section~\ref{sec:curriculum_analysis}). We introduce two variants of KICP: KICP-KA removes the knowledge adaptation module and directly trains the parameters of original LM, and KICP-ATT discards the attribute knowledge and pretrains only on the entity relation knowledge. The results of the two variants are also reported in Table~\ref{tab:overall_results}.~\footnote{The results of KICP-ATT on ComplexWebQuestions, FreebaseQA and Math23K are unavailable, as Wikidata and HowNet do not contain attribute knowledge.} We can summarize the following conclusions. First, the two variants perform worse than KICP, which shows that KA could reduce the negative impacts of generated corpus, and the attribute knowledge is also useful in KBQA. Next, in CN-QA, KICP-ATT performs worse than KICP-KA, which means that attribute knowledge exploitation contributes more than knowledge adaptation on this task. The result is reasonable since a large part of CN-QA requires attribute knowledge (about 45\%). Last, KICP-KA performs worse than BERT in Math23K, which may be due to that KICP-KA hurts the NLU ability of original LM in knowledge pretraining.

\subsubsection{Performance over Difficulty}

\begin{table}
  \caption{Performances on Easy and Hard Questions}
  \label{tab:difficulty_analysis}
  \centering
  \begin{tabular}{c|cc|cc}
    \toprule
    Dataset & \multicolumn{2}{c|}{CN-QA} & \multicolumn{2}{c}{FreebaseQA} \\
    Difficulty & Easy & Hard & Easy & Hard \\
    \midrule
    BERT & 0.633 & 0.603 & 0.920 & 0.891 \\
    KICP & 0.676 & 0.634 & 0.933 & 0.907 \\
    \bottomrule
  \end{tabular}
\end{table}

\begin{figure}[t]
  \centering
  \subfloat[CN-KG]{\includegraphics[width=78px]{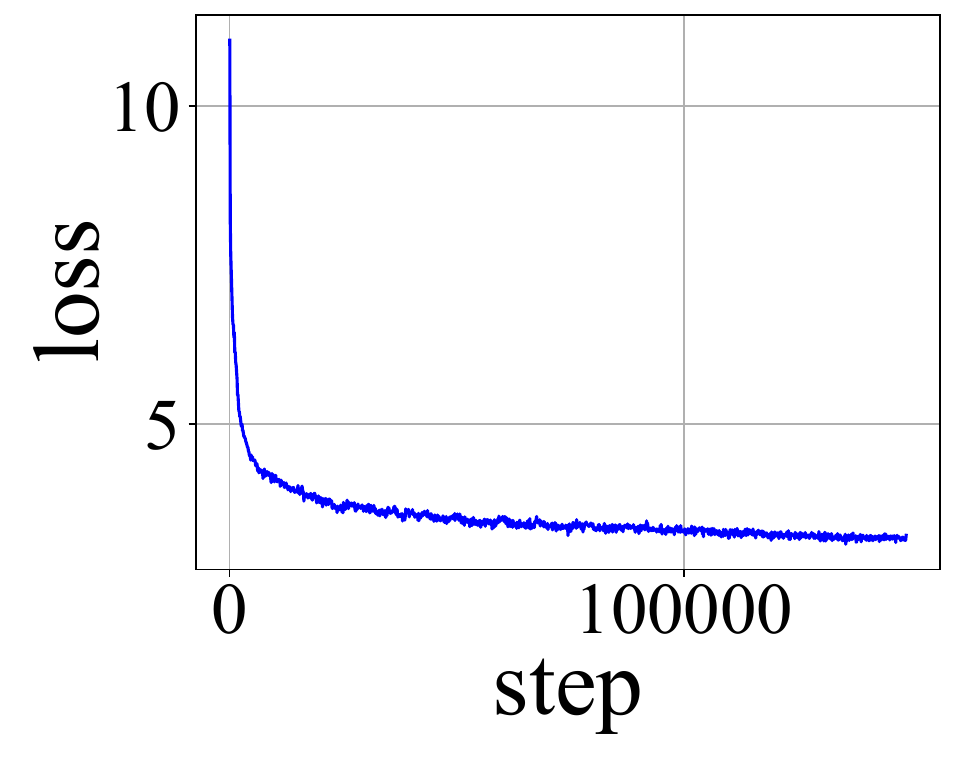}}
  \label{fig:loss_cn_1}
  \hfil
  \subfloat[Wikidata]{\includegraphics[width=78px]{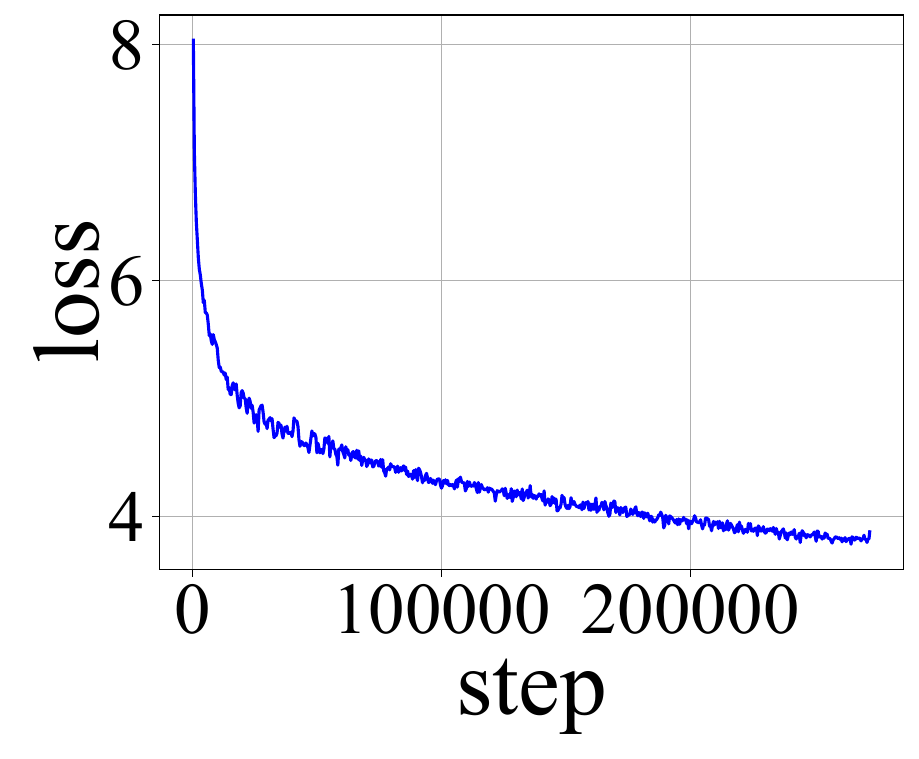}}
  \label{fig:loss_wiki_1}
  \hfil
  \subfloat[HowNet]{\includegraphics[width=78px]{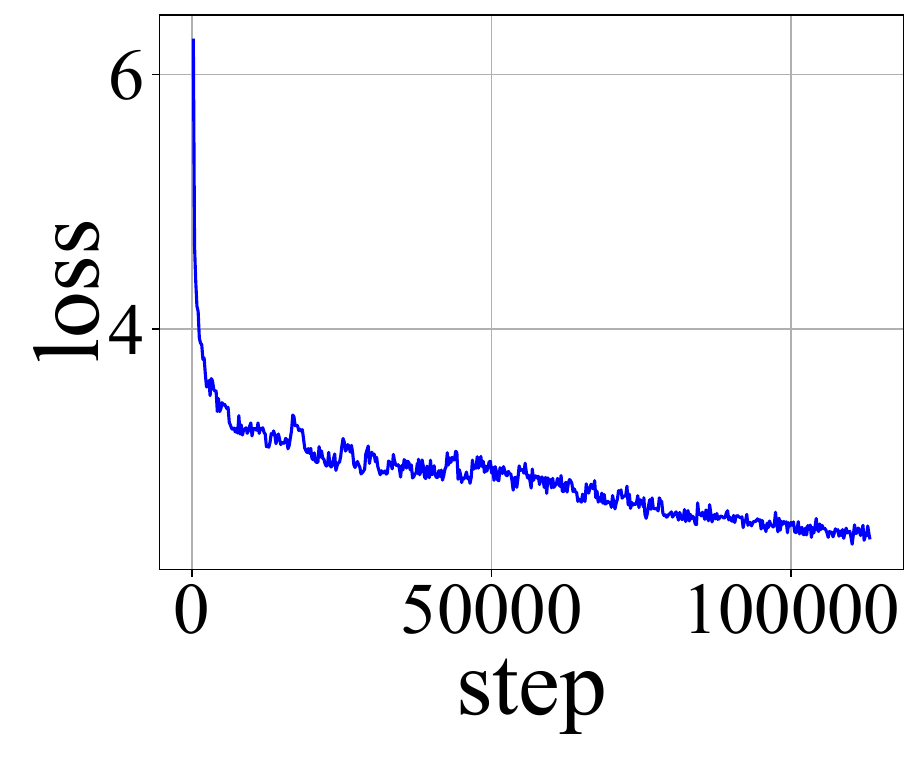}}
  \label{fig:loss_hownet_1}
  \caption{Pretraining loss trend on three KGs in lesson 1.}
  \Description{Pretraining loss trend on three KGs in lesson 1.}
  \label{fig:loss_1}
\end{figure}

We also investigate the performance of KICP on questions with different difficulties to study the complex reasoning ability. We split CN-QA and FreebaseQA into easy questions (answerable with one knowledge triple) and hard ones (requiring multiple triples).~\footnote{ComplexWebQuestions only contains hard questions and Math23K is a generative dataset which is hard to distinguish knowledge requirement, so we do not conduct the experiment on the two datasets.} We report the performances of KICP and BERT in Table~\ref{tab:difficulty_analysis} (F1 on CN-QA and ACC on FreebaseQA). We have the following observations. First, it is reasonable that all methods perform much better on the easy questions than the hard ones. Second, KICP outperforms BERT on both easy and hard questions, showing that both easy and complex QA reasoning benefits from knowledge injection and exploitation. Next, the improvement on hard questions are larger in FreebaseQA. The reason may be that KICP are pretrained on corpus requiring more reasoning ability, which contributes to the higher performance in hard questions. However, in CN-QA the easy questions benefit more, which may result from the much larger proportion of easy questions benefiting from knowledge, and leads to a higher improvement.

\subsection{Curriculum Reasoning Analysis}\label{sec:curriculum_analysis}

In this section, we investigate the feasibility and effectiveness of curriculum reasoning in KICP.

\subsubsection{Loss of Curriculum Pretraining}

Obviously the corpus generated by the CR module greatly differs from the natural ones. Therefore, to verify the feasibility of pretraining with such corpus, we plot the trend of loss in pretraining. Due to limited space, we report the lesson 1 results on three KGs in Figure~\ref{fig:loss_1}.
% ~\footnote{ComplexWebQuestions and FreebaseQA both use Wikidata as KG, so three KGs are included in total in experiments.} 
From the figure, the loss keeps dropping and then gradually converges, which demonstrates that the generated corpus contains enough information to train the LM for knowledge learning, although it may seem odd compared with natural ones.

\begin{figure}[t]
  \centering
  \subfloat[CN-KG]{\includegraphics[width=78px]{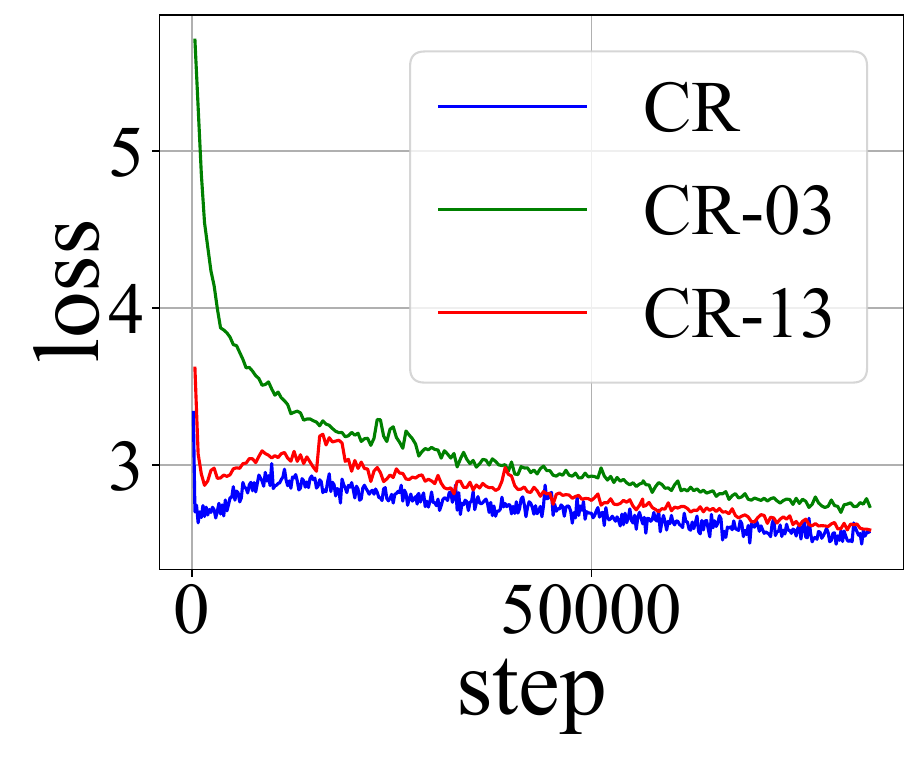}}
  \label{fig:loss_cn_3}
  \hfil
  \subfloat[Wikidata]{\includegraphics[width=78px]{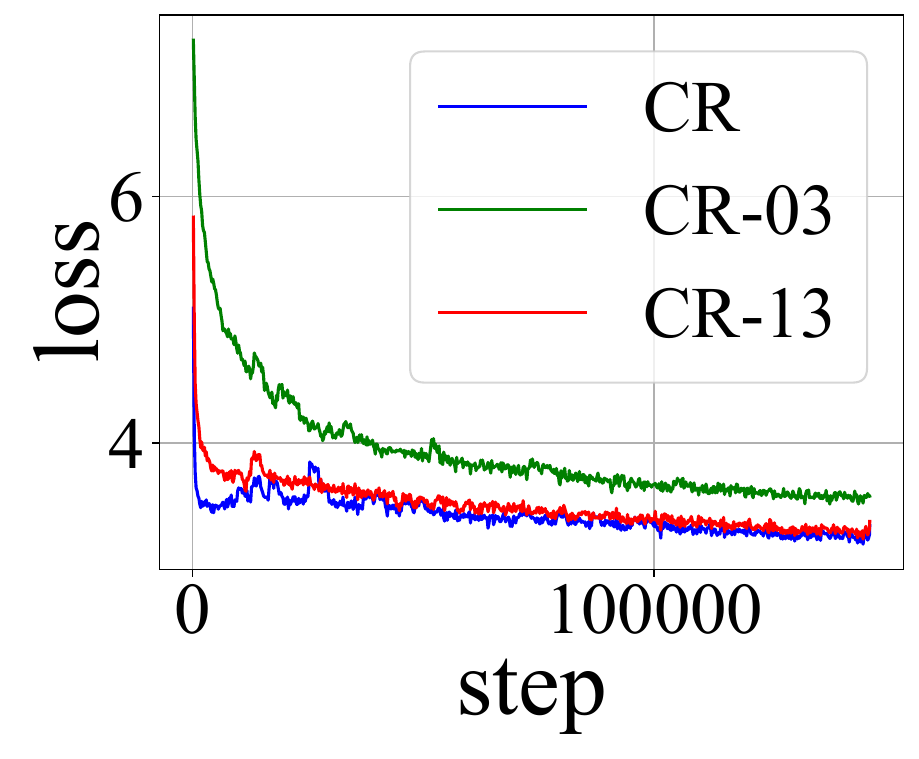}}
  \label{fig:loss_wiki_3}
  \hfil
  \subfloat[HowNet]{\includegraphics[width=78px]{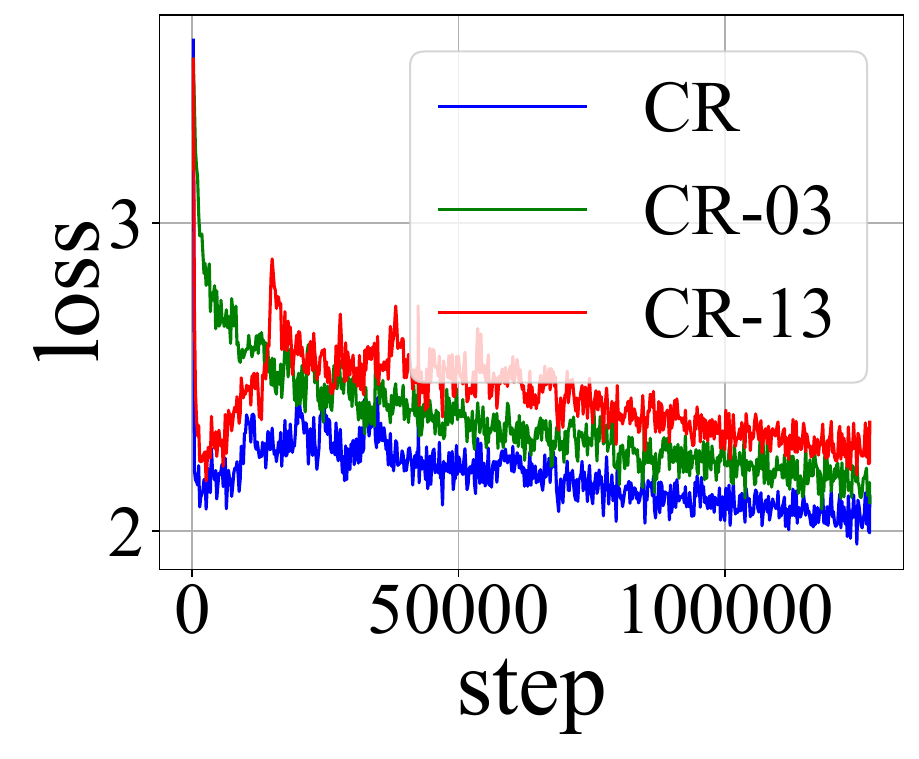}}
  \label{fig:loss_hownet_3}
  \caption{Pretraining loss trend on three KGs in lesson 3.}
  \Description{Pretraining loss trend on three KGs in lesson 3.}
  \label{fig:loss_3}
\end{figure}

\begin{figure}[t]
  \centering
  \subfloat[CN-QA]{\includegraphics[width=95px]{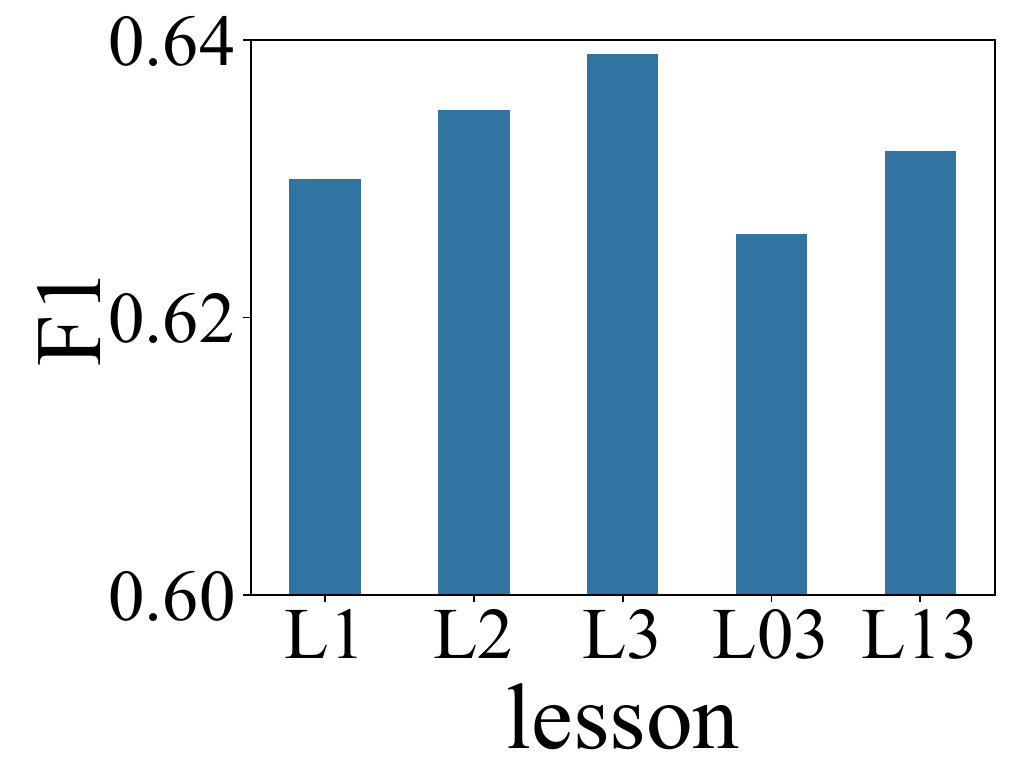}}
  \label{fig:cn_cp}
  \hfil
  \subfloat[ComplexWebQuestions]{\includegraphics[width=95px]{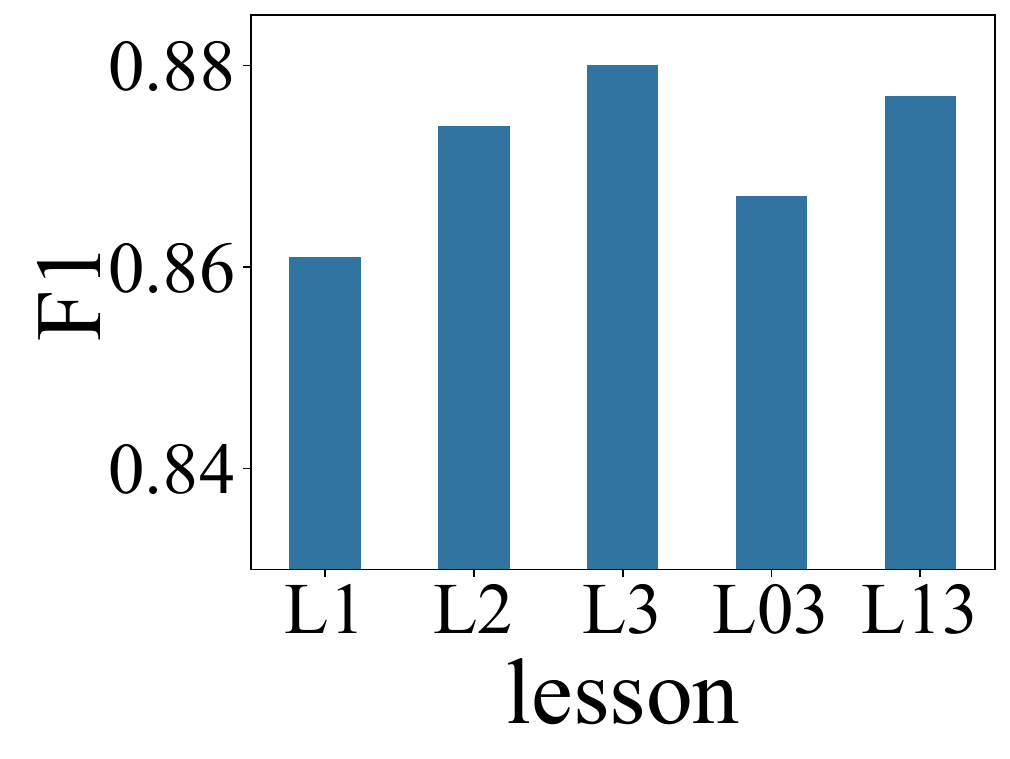}}
  \label{fig:cwq_cp}
  \hfil
  \subfloat[FreebaseQA]{\includegraphics[width=95px]{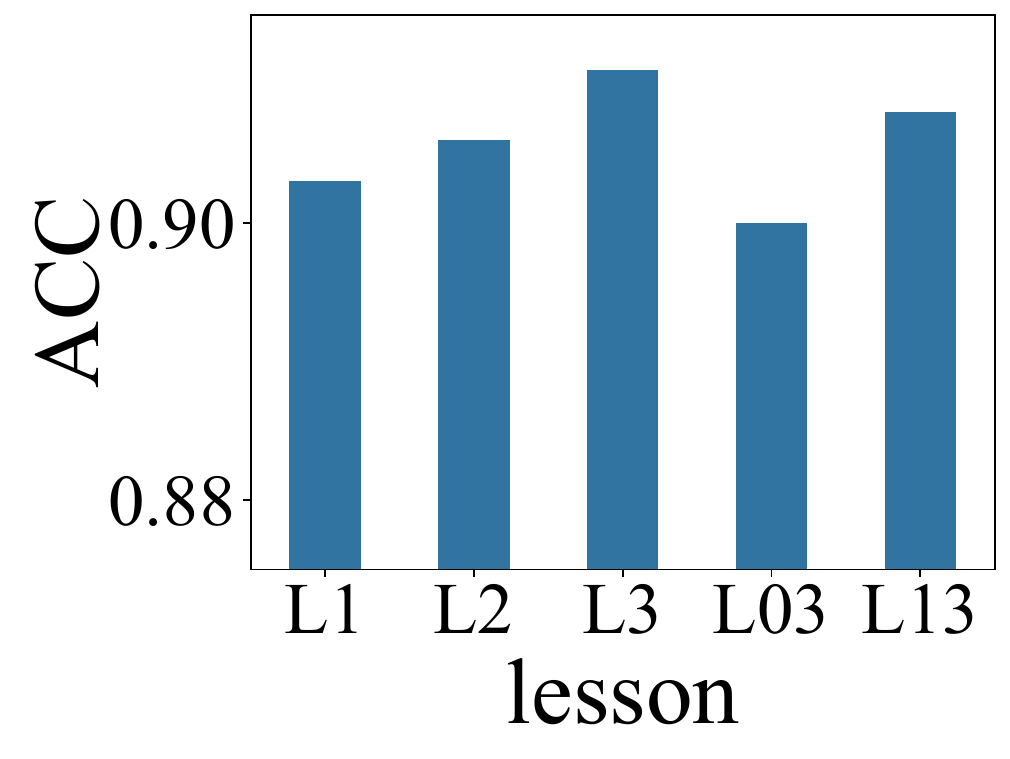}}
  \label{fig:wiki_cp}
  \hfil
  \subfloat[Math23K]{\includegraphics[width=95px]{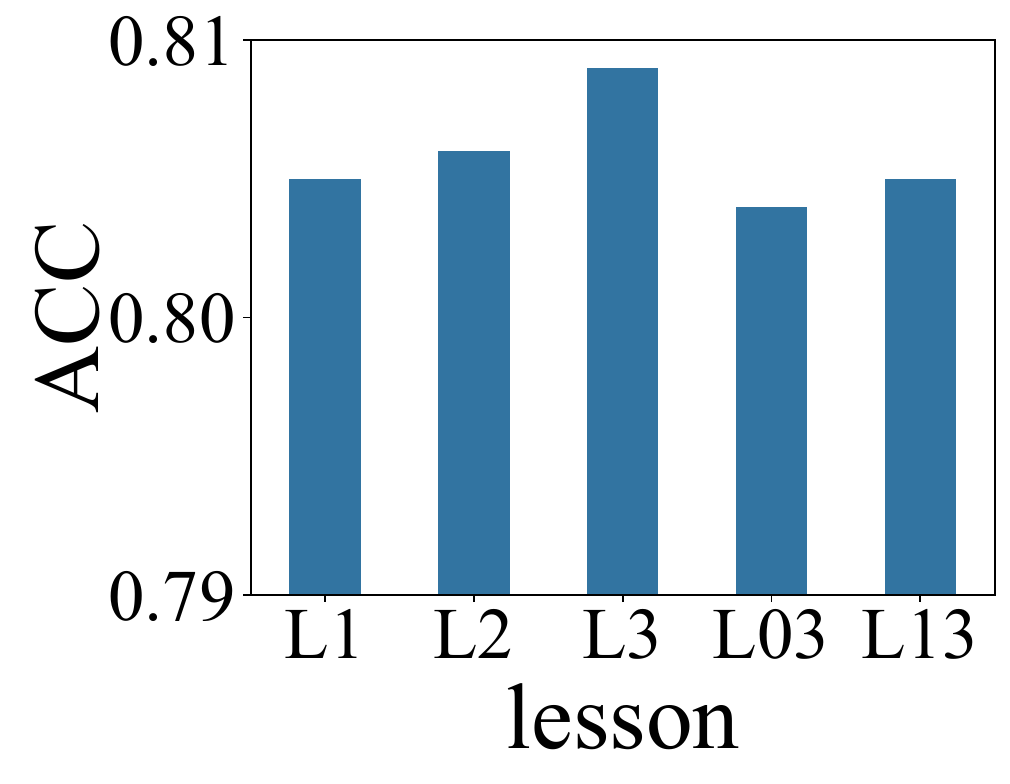}}
  \label{fig:hownet_cp}
  \caption{Performances of LM pretrained for each lesson.}
  \Description{Performances of LM pretrained for each lesson.}
  \label{fig:cp}
\end{figure}

CR also aims to reduce difficulty of pretraining LM for complex reasoning in lesson 3. To investigate the effectiveness, we plot the loss trend in lesson 3 in Figure~\ref{fig:loss_3} with two variants: CR-03 directly trains on lesson 3 without previous lessons, and CR-13 skips lesson 2. There are several observations. First, the loss of CR drops faster and finally reaches lower, proving that the curriculum setting could reduce the training difficulty. Second, the trend of CR-03 is similar to lesson 1 in Figure~\ref{fig:loss_1}, meaning that in CR-03 the model may first learn basic knowledge as lesson 1 and then reasoning. Third, the loss of CR and CR-13 has a short increase in the beginning which may be due to the higher difficulty of lesson 3 and the different distribution from previous easier lesson. Last, CR-13 works better than CR-03 in CN-KG and Wikidata, showing that the LM can perform reasoning better with knowledge memorized. The exception in HowNet may be due to that HowNet mainly contains semantic information, which has been partially covered in LM.

\subsubsection{Performance of Curriculum Reasoning}

We also evaluate the effectiveness of CR on downstream QA tasks. Ideally, the LM performs better after pretrained on each lesson. Therefore, we evaluate the LM finishing lesson 1, 2, 3 (``L1'', ``L2'', ``L3'') with CR-03 and CR-13 (``L03'' and ``L13'') in Figure~\ref{fig:cp}. We can get the following observations. First, performances of models keep increasing after finishing each lesson, which proves the above assumption. Second, L3 performs much better than L03 and L13 (all pretrained on lesson 3), showing that the curriculum setting helps in both convergence and the final outcome. Third, the results can also be viewed as an ablation study on each lesson (``L3'' for ``KICP'', ``L1'' for ``KICP w/o CR'', ``L13'' for ``KICP w/o L2'', ``L2'' for ``KICP w/o L3'', and ``L03'' for ``KICP w/o L1\&L2''), which demonstrates the effectiveness of each lesson. Last, the performances on Math23K do not differ greatly. The reason may be that Math23K requires NLU more than knowledge, thus the effect of pretraining are limited.

\subsection{Training Size Analysis}

\begin{figure}[t]
  \centering
  \subfloat[CN-QA]{\includegraphics[width=95px]{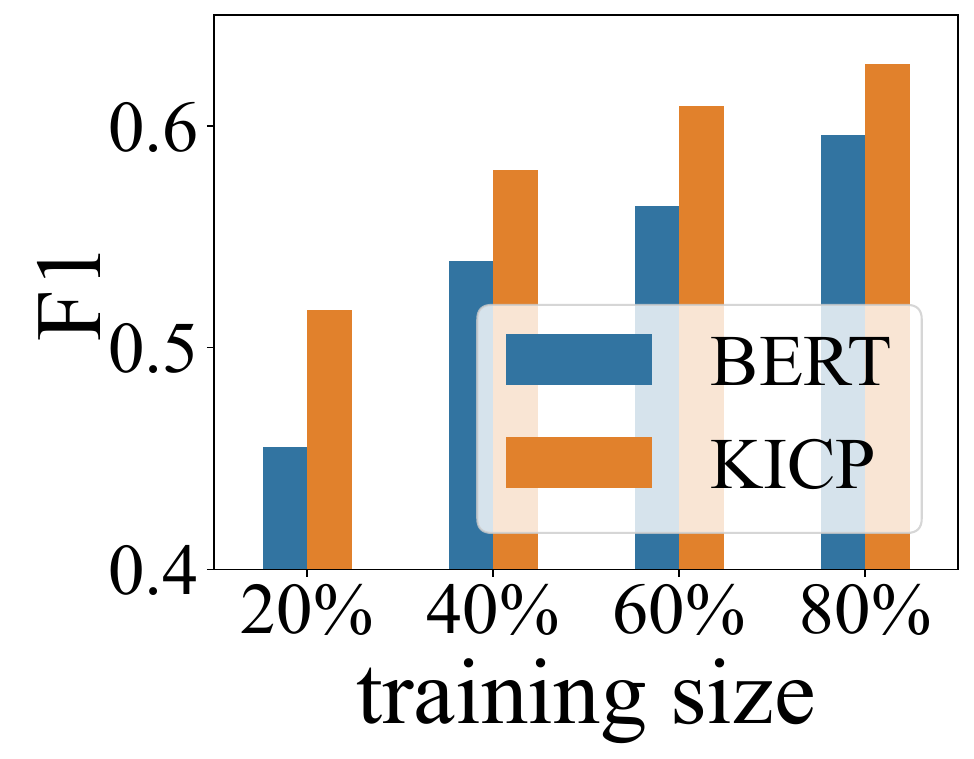}}
  \label{fig:cn_ta}
  \hfil
  \subfloat[ComplexWebQuestions]{\includegraphics[width=95px]{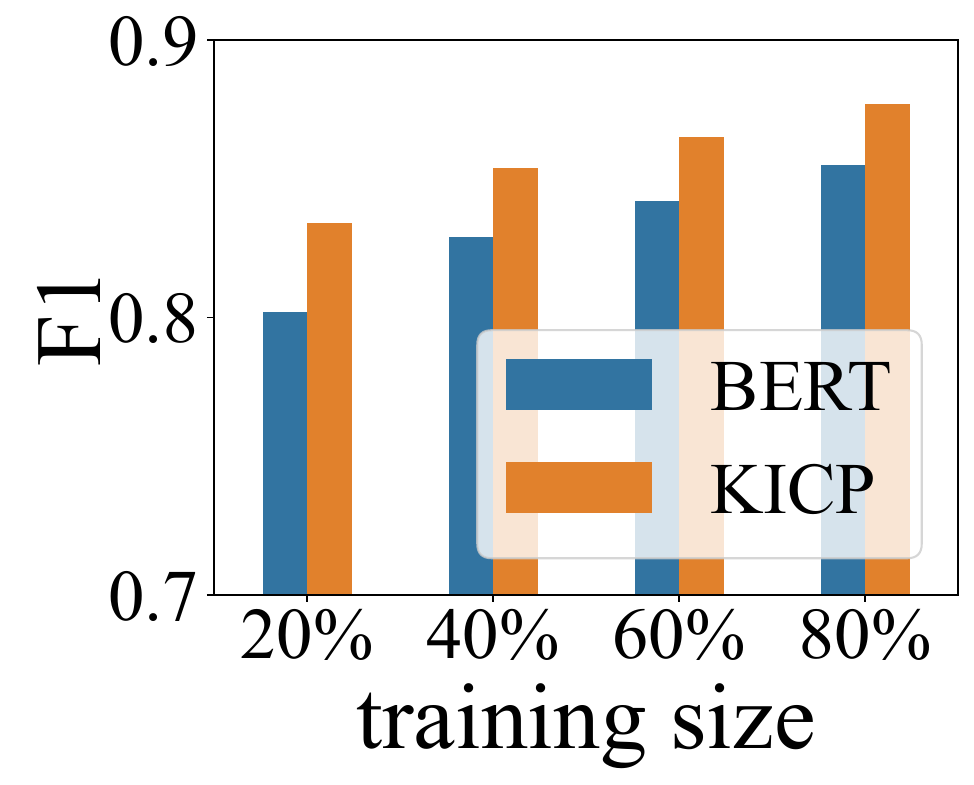}}
  \label{fig:cwq_ta}
  \hfil
  \subfloat[FreebaseQA]{\includegraphics[width=95px]{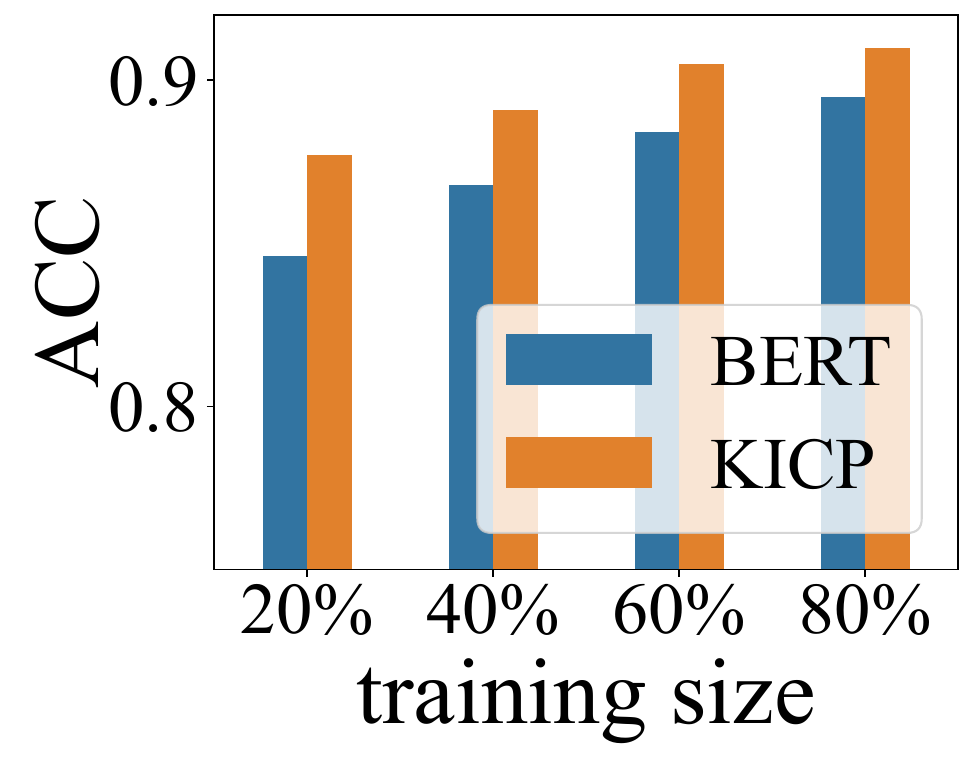}}
  \label{fig:wiki_ta}
  \hfil
  \subfloat[Math23K]{\includegraphics[width=95px]{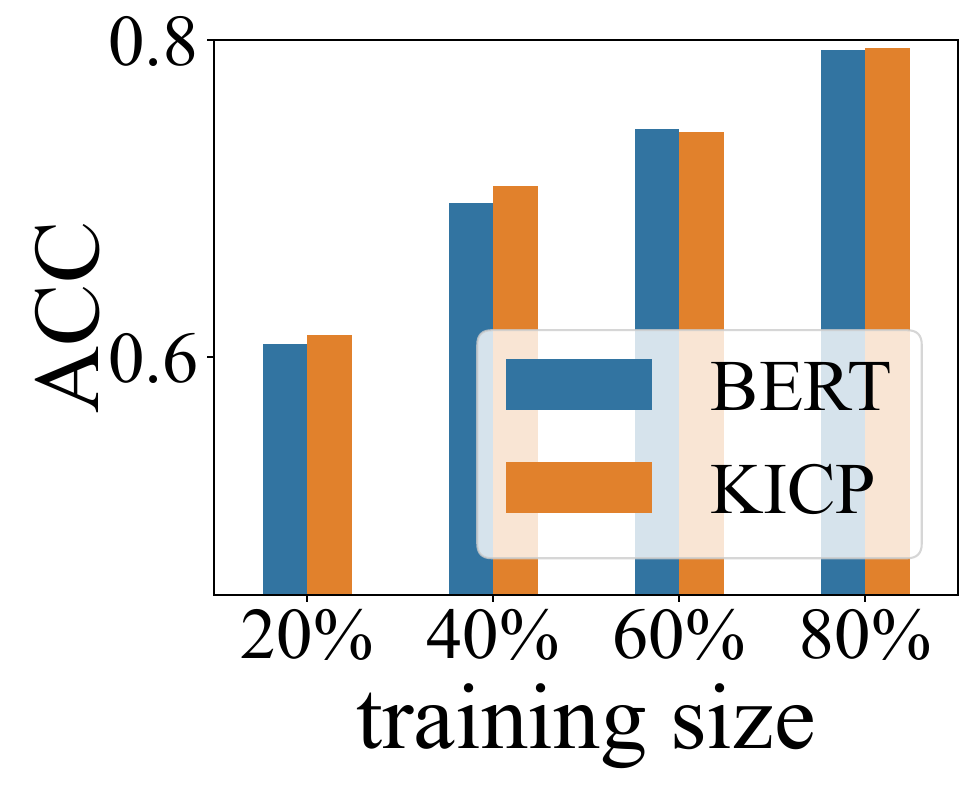}}
  \label{fig:hownet_ta}
  \caption{Performances of KICP and BERT over training size.}
  \Description{Performances of KICP and BERT over training size.}
  \label{fig:ta}
\end{figure}

The pretrained LM aims to reduce the requirement of labeled data and improve the generalization, so the LM pretrained on the KG is expected to have a better performance than the original ones with limited labeled data. Therefore, we split the QA datasets with different training proportion (i.e., 20\%, 40\%, 60\%, 80\%) to evaluate performances of KICP and BERT. The results are demonstrated in Figure~\ref{fig:ta}. From the figure, there are several observations. First, the performances of both KICP and BERT reasonably increase with more training samples. Next, although KICP outperforms BERT in all training settings, generally the differences are larger with less training data. The reason may be that the pretrained KICP could utilize the knowledge learned from KG and exploit less labeled data to learn the mapping from question to answer and achieve a good performance, while BERT needs to learn knowledge from the labeled data, which may be harder without enough data and result in worse performance.

\subsection{Case Study}\label{sec:case_study}

\begin{table}
  \caption{Cases of KICP and BERT}
  \label{tab:case_study}
  \centering
  \begin{tabular}{l}
    \toprule
    \textbf{Case 1:} Who composed the song \textit{Alexander's Ragtime Band} \\ in 1911 ?\\
    \textbf{KICP:} Irving Berlin (\textbf{correct}) \\
    \textbf{BERT:} Woody Guthrie (\textbf{wrong}) \\
    \midrule
    \textbf{Case 2:} Thomas Harris's 1988 novel \textit{The Silence of the Lambs} \\ was actually a sequel - what was the name of the first book in \\ the series ?\\
    \textbf{KICP:} \textit{Red Dragon} (\textbf{correct}) \\
    \textbf{BERT:} \textit{Dubliners} (\textbf{wrong}) \\
    \midrule
    \textbf{Case 3:} Which producer is responsible for \textit{Pearl Harbour}, \\ \textit{Pirates of the Caribbean}, and \textit{Armageddon} ? \\
    \textbf{KICP:} Robert Mulligan (\textbf{wrong}) \\
    \textbf{BERT:} John Ridley (\textbf{wrong}) \\
    \textbf{Answer:} Jerry Bruckheimer \\
    \bottomrule
  \end{tabular}
\end{table}

We demonstrate three typical cases by KICP and BERT on KBQA datasets in Table~\ref{tab:case_study}, and provide more in Appendix~\ref{app:cases}. In case 1, BERT does not understand the knowledge about the lyricist of the song, and fails in the question, while KICP learns related knowledge in pretraining and correctly answer the question. In case 2, KICP is capable of conducting multi-hop reasoning to find the complex relation between ``Thomas Harris'', ``\textit{The Silence of the Lambs}'' and ``\textit{Red Dragon}'' for the answer when the direct relation is unavailable, while BERT does not support such complex reasoning. In case 3, although both methods fail in the question, KICP predicts a closer answer which is also a producer with related knowledge, but BERT fails and makes an unrelated prediction.

\section{Conclusion}

In this paper, we proposed a general Knowledge-Injected Curriculum Pretraining framework (KICP) to learn the KG for question answering, which could work with different detailed techniques for flexible application. We developed a general knowledge injection module to convert the KG into the pretraining corpus for LM with three key steps, and proposed a knowledge adaptation module to reduce the negative impacts of the gap between the generated and natural corpus by keeping the NLU ability of LM in knowledge learning. Furthermore, we designed a curriculum reasoning module to effectively pretrain the LM for human-like complex knowledge reasoning. Experimental results on four QA datasets demonstrated that the proposed KICP could achieve a more comprehensive learning and exploitation of KG for questions answering, and the curriculum setting could effectively reduce the pretraining difficulty and promote the outcome.

The proposed framework still had some limitations. First, the diversity of corpus generated by KICP was limited, and it would benefit if the generated corpus could be more similar to natural ones. Second, in the paper we mainly focused on the LM for language understanding, and we will generalize our framework to generative LM in the future. Last, KICP only exploited the KG as knowledge source, and there were much more types of knowledge to be studied.

\begin{acks}
  This research was partial supported by grants from the National Key Research and Development Program of China (2021YFF0901005) and the National Natural Science Foundation of China (62106244, U20A20229), and the University Synergy Innovation Program of Anhui Province (GXXT-2022-042), and OPPO Research Fund.
\end{acks}

\bibliographystyle{ACM-Reference-Format}
\bibliography{paper}


\begin{thebibliography}{53}

%%% the Web address.

\ifx \showCODEN    \undefined \def \showCODEN     #1{\unskip}     \fi
\ifx \showDOI      \undefined \def \showDOI       #1{#1}\fi
\ifx \showISBNx    \undefined \def \showISBNx     #1{\unskip}     \fi
\ifx \showISBNxiii \undefined \def \showISBNxiii  #1{\unskip}     \fi
\ifx \showISSN     \undefined \def \showISSN      #1{\unskip}     \fi
\ifx \showLCCN     \undefined \def \showLCCN      #1{\unskip}     \fi
\ifx \shownote     \undefined \def \shownote      #1{#1}          \fi
\ifx \showarticletitle \undefined \def \showarticletitle #1{#1}   \fi
\ifx \showURL      \undefined \def \showURL       {\relax}        \fi
% invisible to TeX
\providecommand\bibfield[2]{#2}
\providecommand\bibinfo[2]{#2}
\providecommand\natexlab[1]{#1}
\providecommand\showeprint[2][]{arXiv:#2}

\bibitem[\protect\citeauthoryear{Achiam, Adler, Agarwal, Ahmad, Akkaya, Aleman,
  Almeida, Altenschmidt, Altman, Anadkat, et~al\mbox{.}}{Achiam
  et~al\mbox{.}}{2023}]%
        {achiam2023gpt}
\bibfield{author}{\bibinfo{person}{Josh Achiam}, \bibinfo{person}{Steven
  Adler}, \bibinfo{person}{Sandhini Agarwal}, \bibinfo{person}{Lama Ahmad},
  \bibinfo{person}{Ilge Akkaya}, \bibinfo{person}{Florencia~Leoni Aleman},
  \bibinfo{person}{Diogo Almeida}, \bibinfo{person}{Janko Altenschmidt},
  \bibinfo{person}{Sam Altman}, \bibinfo{person}{Shyamal Anadkat},
  {et~al\mbox{.}}} \bibinfo{year}{2023}\natexlab{}.
\newblock \showarticletitle{Gpt-4 technical report}.
\newblock \bibinfo{journal}{\emph{arXiv preprint arXiv:2303.08774}}
  (\bibinfo{year}{2023}).
\newblock


\bibitem[\protect\citeauthoryear{Agarwal, Ge, Shakeri, and Al-Rfou}{Agarwal
  et~al\mbox{.}}{2021}]%
        {agarwal2021knowledge}
\bibfield{author}{\bibinfo{person}{Oshin Agarwal}, \bibinfo{person}{Heming Ge},
  \bibinfo{person}{Siamak Shakeri}, {and} \bibinfo{person}{Rami Al-Rfou}.}
  \bibinfo{year}{2021}\natexlab{}.
\newblock \showarticletitle{Knowledge Graph Based Synthetic Corpus Generation
  for Knowledge-Enhanced Language Model Pre-training}. In
  \bibinfo{booktitle}{\emph{Proceedings of the 2021 Conference of the North
  American Chapter of the Association for Computational Linguistics: Human
  Language Technologies}}. \bibinfo{pages}{3554--3565}.
\newblock


\bibitem[\protect\citeauthoryear{Chen, Su, Yan, and Wang}{Chen
  et~al\mbox{.}}{2020}]%
        {chen2020kgpt}
\bibfield{author}{\bibinfo{person}{Wenhu Chen}, \bibinfo{person}{Yu Su},
  \bibinfo{person}{Xifeng Yan}, {and} \bibinfo{person}{William~Yang Wang}.}
  \bibinfo{year}{2020}\natexlab{}.
\newblock \showarticletitle{KGPT: Knowledge-Grounded Pre-Training for
  Data-to-Text Generation}. In \bibinfo{booktitle}{\emph{Proceedings of the
  2020 Conference on Empirical Methods in Natural Language Processing
  (EMNLP)}}. \bibinfo{pages}{8635--8648}.
\newblock


\bibitem[\protect\citeauthoryear{Chen, Zhang, Xie, Deng, Yao, Tan, Huang, Si,
  and Chen}{Chen et~al\mbox{.}}{2022}]%
        {chen2022knowprompt}
\bibfield{author}{\bibinfo{person}{Xiang Chen}, \bibinfo{person}{Ningyu Zhang},
  \bibinfo{person}{Xin Xie}, \bibinfo{person}{Shumin Deng},
  \bibinfo{person}{Yunzhi Yao}, \bibinfo{person}{Chuanqi Tan},
  \bibinfo{person}{Fei Huang}, \bibinfo{person}{Luo Si}, {and}
  \bibinfo{person}{Huajun Chen}.} \bibinfo{year}{2022}\natexlab{}.
\newblock \showarticletitle{Knowprompt: Knowledge-aware prompt-tuning with
  synergistic optimization for relation extraction}. In
  \bibinfo{booktitle}{\emph{Proceedings of the ACM Web conference 2022}}.
  \bibinfo{pages}{2778--2788}.
\newblock


\bibitem[\protect\citeauthoryear{Cui, Che, Liu, Qin, and Yang}{Cui
  et~al\mbox{.}}{2021}]%
        {cui2021pre}
\bibfield{author}{\bibinfo{person}{Yiming Cui}, \bibinfo{person}{Wanxiang Che},
  \bibinfo{person}{Ting Liu}, \bibinfo{person}{Bing Qin}, {and}
  \bibinfo{person}{Ziqing Yang}.} \bibinfo{year}{2021}\natexlab{}.
\newblock \showarticletitle{Pre-training with whole word masking for chinese
  bert}.
\newblock \bibinfo{journal}{\emph{IEEE/ACM Transactions on Audio, Speech, and
  Language Processing}}  \bibinfo{volume}{29} (\bibinfo{year}{2021}),
  \bibinfo{pages}{3504--3514}.
\newblock


\bibitem[\protect\citeauthoryear{Devlin, Chang, Lee, and Toutanova}{Devlin
  et~al\mbox{.}}{2019}]%
        {devlin2019bert}
\bibfield{author}{\bibinfo{person}{Jacob Devlin}, \bibinfo{person}{Ming-Wei
  Chang}, \bibinfo{person}{Kenton Lee}, {and} \bibinfo{person}{Kristina
  Toutanova}.} \bibinfo{year}{2019}\natexlab{}.
\newblock \showarticletitle{BERT: Pre-training of Deep Bidirectional
  Transformers for Language Understanding}. In
  \bibinfo{booktitle}{\emph{Proceedings of the 2019 Conference of the North
  American Chapter of the Association for Computational Linguistics: Human
  Language Technologies, Volume 1 (Long and Short Papers)}}.
  \bibinfo{pages}{4171--4186}.
\newblock


\bibitem[\protect\citeauthoryear{Ding, Qin, Yang, Wei, Yang, Su, Hu, Chen,
  Chan, Chen, et~al\mbox{.}}{Ding et~al\mbox{.}}{2022}]%
        {ding2022delta}
\bibfield{author}{\bibinfo{person}{Ning Ding}, \bibinfo{person}{Yujia Qin},
  \bibinfo{person}{Guang Yang}, \bibinfo{person}{Fuchao Wei},
  \bibinfo{person}{Zonghan Yang}, \bibinfo{person}{Yusheng Su},
  \bibinfo{person}{Shengding Hu}, \bibinfo{person}{Yulin Chen},
  \bibinfo{person}{Chi-Min Chan}, \bibinfo{person}{Weize Chen},
  {et~al\mbox{.}}} \bibinfo{year}{2022}\natexlab{}.
\newblock \showarticletitle{Delta tuning: A comprehensive study of parameter
  efficient methods for pre-trained language models}.
\newblock \bibinfo{journal}{\emph{arXiv preprint arXiv:2203.06904}}
  (\bibinfo{year}{2022}).
\newblock


\bibitem[\protect\citeauthoryear{Du, Qian, Liu, Ding, Qiu, Yang, and Tang}{Du
  et~al\mbox{.}}{2022}]%
        {du2022glm}
\bibfield{author}{\bibinfo{person}{Zhengxiao Du}, \bibinfo{person}{Yujie Qian},
  \bibinfo{person}{Xiao Liu}, \bibinfo{person}{Ming Ding},
  \bibinfo{person}{Jiezhong Qiu}, \bibinfo{person}{Zhilin Yang}, {and}
  \bibinfo{person}{Jie Tang}.} \bibinfo{year}{2022}\natexlab{}.
\newblock \showarticletitle{GLM: General Language Model Pretraining with
  Autoregressive Blank Infilling}. In \bibinfo{booktitle}{\emph{Proceedings of
  the 60th Annual Meeting of the Association for Computational Linguistics
  (Volume 1: Long Papers)}}. \bibinfo{pages}{320--335}.
\newblock


\bibitem[\protect\citeauthoryear{Feng, Balachandran, Bai, and Tsvetkov}{Feng
  et~al\mbox{.}}{2023}]%
        {feng2023factkb}
\bibfield{author}{\bibinfo{person}{Shangbin Feng}, \bibinfo{person}{Vidhisha
  Balachandran}, \bibinfo{person}{Yuyang Bai}, {and} \bibinfo{person}{Yulia
  Tsvetkov}.} \bibinfo{year}{2023}\natexlab{}.
\newblock \showarticletitle{Factkb: Generalizable factuality evaluation using
  language models enhanced with factual knowledge}.
\newblock \bibinfo{journal}{\emph{arXiv preprint arXiv:2305.08281}}
  (\bibinfo{year}{2023}).
\newblock


\bibitem[\protect\citeauthoryear{Hu, Xu, Yu, Wang, Yang, Zhu, Chang, and
  Sun}{Hu et~al\mbox{.}}{2022}]%
        {hu2022empowering}
\bibfield{author}{\bibinfo{person}{Ziniu Hu}, \bibinfo{person}{Yichong Xu},
  \bibinfo{person}{Wenhao Yu}, \bibinfo{person}{Shuohang Wang},
  \bibinfo{person}{Ziyi Yang}, \bibinfo{person}{Chenguang Zhu},
  \bibinfo{person}{Kai-Wei Chang}, {and} \bibinfo{person}{Yizhou Sun}.}
  \bibinfo{year}{2022}\natexlab{}.
\newblock \showarticletitle{Empowering language models with knowledge graph
  reasoning for open-domain question answering}. In
  \bibinfo{booktitle}{\emph{Proceedings of the 2022 Conference on Empirical
  Methods in Natural Language Processing}}. \bibinfo{pages}{9562--9581}.
\newblock


\bibitem[\protect\citeauthoryear{Huang, Zhang, Li, and Li}{Huang
  et~al\mbox{.}}{2019}]%
        {huang2019knowledge}
\bibfield{author}{\bibinfo{person}{Xiao Huang}, \bibinfo{person}{Jingyuan
  Zhang}, \bibinfo{person}{Dingcheng Li}, {and} \bibinfo{person}{Ping Li}.}
  \bibinfo{year}{2019}\natexlab{}.
\newblock \showarticletitle{Knowledge graph embedding based question
  answering}. In \bibinfo{booktitle}{\emph{Proceedings of the twelfth ACM
  international conference on web search and data mining}}.
  \bibinfo{pages}{105--113}.
\newblock


\bibitem[\protect\citeauthoryear{Huang, Lin, Wang, Liu, Chen, Ma, Su, and
  Tong}{Huang et~al\mbox{.}}{2021}]%
        {huang2021disenqnet}
\bibfield{author}{\bibinfo{person}{Zhenya Huang}, \bibinfo{person}{Xin Lin},
  \bibinfo{person}{Hao Wang}, \bibinfo{person}{Qi Liu}, \bibinfo{person}{Enhong
  Chen}, \bibinfo{person}{Jianhui Ma}, \bibinfo{person}{Yu Su}, {and}
  \bibinfo{person}{Wei Tong}.} \bibinfo{year}{2021}\natexlab{}.
\newblock \showarticletitle{Disenqnet: Disentangled representation learning for
  educational questions}. In \bibinfo{booktitle}{\emph{Proceedings of the 27th
  ACM SIGKDD Conference on Knowledge Discovery \& Data Mining}}.
  \bibinfo{pages}{696--704}.
\newblock


\bibitem[\protect\citeauthoryear{Jiang, Wu, and Jiang}{Jiang
  et~al\mbox{.}}{2019}]%
        {jiang2019freebaseqa}
\bibfield{author}{\bibinfo{person}{Kelvin Jiang}, \bibinfo{person}{Dekun Wu},
  {and} \bibinfo{person}{Hui Jiang}.} \bibinfo{year}{2019}\natexlab{}.
\newblock \showarticletitle{FreebaseQA: A New Factoid QA Data Set Matching
  Trivia-Style Question-Answer Pairs with Freebase}. In
  \bibinfo{booktitle}{\emph{Proceedings of NAACL-HLT}}.
  \bibinfo{pages}{318--323}.
\newblock


\bibitem[\protect\citeauthoryear{Lin, Chen, Chen, and Ren}{Lin
  et~al\mbox{.}}{2019}]%
        {lin2019kagnet}
\bibfield{author}{\bibinfo{person}{Bill~Yuchen Lin}, \bibinfo{person}{Xinyue
  Chen}, \bibinfo{person}{Jamin Chen}, {and} \bibinfo{person}{Xiang Ren}.}
  \bibinfo{year}{2019}\natexlab{}.
\newblock \showarticletitle{KagNet: Knowledge-Aware Graph Networks for
  Commonsense Reasoning}. In \bibinfo{booktitle}{\emph{Proceedings of the 2019
  Conference on Empirical Methods in Natural Language Processing and the 9th
  International Joint Conference on Natural Language Processing
  (EMNLP-IJCNLP)}}. \bibinfo{pages}{2829--2839}.
\newblock


\bibitem[\protect\citeauthoryear{Lin, Huang, Zhao, Chen, Liu, Lian, Li, and
  Wang}{Lin et~al\mbox{.}}{2023}]%
        {lin2023learning}
\bibfield{author}{\bibinfo{person}{Xin Lin}, \bibinfo{person}{Zhenya Huang},
  \bibinfo{person}{Hongke Zhao}, \bibinfo{person}{Enhong Chen},
  \bibinfo{person}{Qi Liu}, \bibinfo{person}{Defu Lian}, \bibinfo{person}{Xin
  Li}, {and} \bibinfo{person}{Hao Wang}.} \bibinfo{year}{2023}\natexlab{}.
\newblock \showarticletitle{Learning Relation-Enhanced Hierarchical Solver for
  Math Word Problems}.
\newblock \bibinfo{journal}{\emph{IEEE Transactions on Neural Networks and
  Learning Systems}} (\bibinfo{year}{2023}).
\newblock


\bibitem[\protect\citeauthoryear{Liu, Huang, Lin, Liu, Ma, and Chen}{Liu
  et~al\mbox{.}}{2022a}]%
        {liu2022cognitive}
\bibfield{author}{\bibinfo{person}{Jiayu Liu}, \bibinfo{person}{Zhenya Huang},
  \bibinfo{person}{Xin Lin}, \bibinfo{person}{Qi Liu}, \bibinfo{person}{Jianhui
  Ma}, {and} \bibinfo{person}{Enhong Chen}.} \bibinfo{year}{2022}\natexlab{a}.
\newblock \showarticletitle{A cognitive solver with autonomously knowledge
  learning for reasoning mathematical answers}. In
  \bibinfo{booktitle}{\emph{2022 IEEE International Conference on Data Mining
  (ICDM)}}. IEEE, \bibinfo{pages}{269--278}.
\newblock


\bibitem[\protect\citeauthoryear{Liu, Huang, Ma, Liu, Chen, Su, and Liu}{Liu
  et~al\mbox{.}}{2023b}]%
        {liu2023guiding}
\bibfield{author}{\bibinfo{person}{Jiayu Liu}, \bibinfo{person}{Zhenya Huang},
  \bibinfo{person}{Zhiyuan Ma}, \bibinfo{person}{Qi Liu},
  \bibinfo{person}{Enhong Chen}, \bibinfo{person}{Tianhuang Su}, {and}
  \bibinfo{person}{Haifeng Liu}.} \bibinfo{year}{2023}\natexlab{b}.
\newblock \showarticletitle{Guiding Mathematical Reasoning via Mastering
  Commonsense Formula Knowledge}. In \bibinfo{booktitle}{\emph{Proceedings of
  the 29th ACM SIGKDD Conference on Knowledge Discovery and Data Mining}}.
  \bibinfo{pages}{1477--1488}.
\newblock


\bibitem[\protect\citeauthoryear{Liu, Huang, Zhai, and Liu}{Liu
  et~al\mbox{.}}{2023c}]%
        {liu2023learning}
\bibfield{author}{\bibinfo{person}{Jiayu Liu}, \bibinfo{person}{Zhenya Huang},
  \bibinfo{person}{Chengxiang Zhai}, {and} \bibinfo{person}{Qi Liu}.}
  \bibinfo{year}{2023}\natexlab{c}.
\newblock \showarticletitle{Learning by Applying: A General Framework for
  Mathematical Reasoning via Enhancing Explicit Knowledge Learning}.
\newblock \bibinfo{journal}{\emph{arXiv preprint arXiv:2302.05717}}
  (\bibinfo{year}{2023}).
\newblock


\bibitem[\protect\citeauthoryear{Liu, Chen, Das, Yang, and Tong}{Liu
  et~al\mbox{.}}{2023a}]%
        {liu2023knowledge}
\bibfield{author}{\bibinfo{person}{Lihui Liu}, \bibinfo{person}{Yuzhong Chen},
  \bibinfo{person}{Mahashweta Das}, \bibinfo{person}{Hao Yang}, {and}
  \bibinfo{person}{Hanghang Tong}.} \bibinfo{year}{2023}\natexlab{a}.
\newblock \showarticletitle{Knowledge Graph Question Answering with Ambiguous
  Query}. In \bibinfo{booktitle}{\emph{Proceedings of the ACM Web Conference
  2023}}. \bibinfo{pages}{2477--2486}.
\newblock


\bibitem[\protect\citeauthoryear{Liu, Li, He, Bing, Joty, and Si}{Liu
  et~al\mbox{.}}{2022b}]%
        {liu2022enhancing}
\bibfield{author}{\bibinfo{person}{Linlin Liu}, \bibinfo{person}{Xin Li},
  \bibinfo{person}{Ruidan He}, \bibinfo{person}{Lidong Bing},
  \bibinfo{person}{Shafiq Joty}, {and} \bibinfo{person}{Luo Si}.}
  \bibinfo{year}{2022}\natexlab{b}.
\newblock \showarticletitle{Enhancing multilingual language model with massive
  multilingual knowledge triples}. In \bibinfo{booktitle}{\emph{Proceedings of
  the 2022 Conference on Empirical Methods in Natural Language Processing}}.
  \bibinfo{pages}{6878--6890}.
\newblock


\bibitem[\protect\citeauthoryear{Liu, Zhou, Zhao, Wang, Ju, Deng, and Wang}{Liu
  et~al\mbox{.}}{2020}]%
        {liu2020k}
\bibfield{author}{\bibinfo{person}{Weijie Liu}, \bibinfo{person}{Peng Zhou},
  \bibinfo{person}{Zhe Zhao}, \bibinfo{person}{Zhiruo Wang},
  \bibinfo{person}{Qi Ju}, \bibinfo{person}{Haotang Deng}, {and}
  \bibinfo{person}{Ping Wang}.} \bibinfo{year}{2020}\natexlab{}.
\newblock \showarticletitle{K-bert: Enabling language representation with
  knowledge graph}. In \bibinfo{booktitle}{\emph{Proceedings of the AAAI
  Conference on Artificial Intelligence}}, Vol.~\bibinfo{volume}{34}.
  \bibinfo{pages}{2901--2908}.
\newblock


\bibitem[\protect\citeauthoryear{Liu, Ott, Goyal, Du, Joshi, Chen, Levy, Lewis,
  Zettlemoyer, and Stoyanov}{Liu et~al\mbox{.}}{2019}]%
        {liu2019roberta}
\bibfield{author}{\bibinfo{person}{Yinhan Liu}, \bibinfo{person}{Myle Ott},
  \bibinfo{person}{Naman Goyal}, \bibinfo{person}{Jingfei Du},
  \bibinfo{person}{Mandar Joshi}, \bibinfo{person}{Danqi Chen},
  \bibinfo{person}{Omer Levy}, \bibinfo{person}{Mike Lewis},
  \bibinfo{person}{Luke Zettlemoyer}, {and} \bibinfo{person}{Veselin
  Stoyanov}.} \bibinfo{year}{2019}\natexlab{}.
\newblock \showarticletitle{Roberta: A robustly optimized bert pretraining
  approach}.
\newblock \bibinfo{journal}{\emph{arXiv preprint arXiv:1907.11692}}
  (\bibinfo{year}{2019}).
\newblock


\bibitem[\protect\citeauthoryear{Logan, Liu, Peters, Gardner, and Singh}{Logan
  et~al\mbox{.}}{2019}]%
        {logan2019barack}
\bibfield{author}{\bibinfo{person}{Robert Logan}, \bibinfo{person}{Nelson~F
  Liu}, \bibinfo{person}{Matthew~E Peters}, \bibinfo{person}{Matt Gardner},
  {and} \bibinfo{person}{Sameer Singh}.} \bibinfo{year}{2019}\natexlab{}.
\newblock \showarticletitle{Barack's Wife Hillary: Using Knowledge Graphs for
  Fact-Aware Language Modeling}. In \bibinfo{booktitle}{\emph{Proceedings of
  the 57th Annual Meeting of the Association for Computational Linguistics}}.
  \bibinfo{pages}{5962--5971}.
\newblock


\bibitem[\protect\citeauthoryear{Loshchilov and Hutter}{Loshchilov and
  Hutter}{2019}]%
        {loshchilovdecoupled}
\bibfield{author}{\bibinfo{person}{Ilya Loshchilov} {and}
  \bibinfo{person}{Frank Hutter}.} \bibinfo{year}{2019}\natexlab{}.
\newblock \showarticletitle{Decoupled Weight Decay Regularization}. In
  \bibinfo{booktitle}{\emph{International Conference on Learning
  Representations}}.
\newblock


\bibitem[\protect\citeauthoryear{Lu, Qiu, Chang, Wu, Zhu, Rajpurohit, Clark,
  and Kalyan}{Lu et~al\mbox{.}}{2022b}]%
        {lu2022dynamic}
\bibfield{author}{\bibinfo{person}{Pan Lu}, \bibinfo{person}{Liang Qiu},
  \bibinfo{person}{Kai-Wei Chang}, \bibinfo{person}{Ying~Nian Wu},
  \bibinfo{person}{Song-Chun Zhu}, \bibinfo{person}{Tanmay Rajpurohit},
  \bibinfo{person}{Peter Clark}, {and} \bibinfo{person}{Ashwin Kalyan}.}
  \bibinfo{year}{2022}\natexlab{b}.
\newblock \showarticletitle{Dynamic prompt learning via policy gradient for
  semi-structured mathematical reasoning}.
\newblock \bibinfo{journal}{\emph{arXiv preprint arXiv:2209.14610}}
  (\bibinfo{year}{2022}).
\newblock


\bibitem[\protect\citeauthoryear{Lu, Lu, Fu, and Liu}{Lu
  et~al\mbox{.}}{2022a}]%
        {lu2022kelm}
\bibfield{author}{\bibinfo{person}{Yinquan Lu}, \bibinfo{person}{Haonan Lu},
  \bibinfo{person}{Guirong Fu}, {and} \bibinfo{person}{Qun Liu}.}
  \bibinfo{year}{2022}\natexlab{a}.
\newblock \showarticletitle{KELM: Knowledge Enhanced Pre-Trained Language
  Representations with Message Passing on Hierarchical Relational Graphs}. In
  \bibinfo{booktitle}{\emph{ICLR 2022 Workshop on Deep Learning on Graphs for
  Natural Language Processing}}.
\newblock


\bibitem[\protect\citeauthoryear{Lukovnikov, Fischer, and Lehmann}{Lukovnikov
  et~al\mbox{.}}{2019}]%
        {lukovnikov2019pretrained}
\bibfield{author}{\bibinfo{person}{Denis Lukovnikov}, \bibinfo{person}{Asja
  Fischer}, {and} \bibinfo{person}{Jens Lehmann}.}
  \bibinfo{year}{2019}\natexlab{}.
\newblock \showarticletitle{Pretrained Transformers for Simple Question
  Answering over Knowledge Graphs}. In \bibinfo{booktitle}{\emph{The Semantic
  Web--ISWC 2019: 18th International Semantic Web Conference, Auckland, New
  Zealand, October 26--30, 2019, Proceedings, Part I}}.
  \bibinfo{pages}{470--486}.
\newblock


\bibitem[\protect\citeauthoryear{Lv, Guo, Xu, Tang, Duan, Gong, Shou, Jiang,
  Cao, and Hu}{Lv et~al\mbox{.}}{2020}]%
        {lv2020graph}
\bibfield{author}{\bibinfo{person}{Shangwen Lv}, \bibinfo{person}{Daya Guo},
  \bibinfo{person}{Jingjing Xu}, \bibinfo{person}{Duyu Tang},
  \bibinfo{person}{Nan Duan}, \bibinfo{person}{Ming Gong},
  \bibinfo{person}{Linjun Shou}, \bibinfo{person}{Daxin Jiang},
  \bibinfo{person}{Guihong Cao}, {and} \bibinfo{person}{Songlin Hu}.}
  \bibinfo{year}{2020}\natexlab{}.
\newblock \showarticletitle{Graph-based reasoning over heterogeneous external
  knowledge for commonsense question answering}. In
  \bibinfo{booktitle}{\emph{Proceedings of the AAAI conference on artificial
  intelligence}}, Vol.~\bibinfo{volume}{34}. \bibinfo{pages}{8449--8456}.
\newblock


\bibitem[\protect\citeauthoryear{Meng, Zhang, Huang, Zhang, and Han}{Meng
  et~al\mbox{.}}{2022}]%
        {meng2022topic}
\bibfield{author}{\bibinfo{person}{Yu Meng}, \bibinfo{person}{Yunyi Zhang},
  \bibinfo{person}{Jiaxin Huang}, \bibinfo{person}{Yu Zhang}, {and}
  \bibinfo{person}{Jiawei Han}.} \bibinfo{year}{2022}\natexlab{}.
\newblock \showarticletitle{Topic discovery via latent space clustering of
  pretrained language model representations}. In
  \bibinfo{booktitle}{\emph{Proceedings of the ACM Web Conference 2022}}.
  \bibinfo{pages}{3143--3152}.
\newblock


\bibitem[\protect\citeauthoryear{Ouyang, Wu, Jiang, Almeida, Wainwright,
  Mishkin, Zhang, Agarwal, Slama, Ray, et~al\mbox{.}}{Ouyang
  et~al\mbox{.}}{2022}]%
        {ouyang2022training}
\bibfield{author}{\bibinfo{person}{Long Ouyang}, \bibinfo{person}{Jeffrey Wu},
  \bibinfo{person}{Xu Jiang}, \bibinfo{person}{Diogo Almeida},
  \bibinfo{person}{Carroll Wainwright}, \bibinfo{person}{Pamela Mishkin},
  \bibinfo{person}{Chong Zhang}, \bibinfo{person}{Sandhini Agarwal},
  \bibinfo{person}{Katarina Slama}, \bibinfo{person}{Alex Ray},
  {et~al\mbox{.}}} \bibinfo{year}{2022}\natexlab{}.
\newblock \showarticletitle{Training language models to follow instructions
  with human feedback}.
\newblock \bibinfo{journal}{\emph{Advances in Neural Information Processing
  Systems}}  \bibinfo{volume}{35} (\bibinfo{year}{2022}),
  \bibinfo{pages}{27730--27744}.
\newblock


\bibitem[\protect\citeauthoryear{Peters, Neumann, Logan, Schwartz, Joshi,
  Singh, and Smith}{Peters et~al\mbox{.}}{2019}]%
        {peters2019knowledge}
\bibfield{author}{\bibinfo{person}{Matthew~E Peters}, \bibinfo{person}{Mark
  Neumann}, \bibinfo{person}{Robert Logan}, \bibinfo{person}{Roy Schwartz},
  \bibinfo{person}{Vidur Joshi}, \bibinfo{person}{Sameer Singh}, {and}
  \bibinfo{person}{Noah~A Smith}.} \bibinfo{year}{2019}\natexlab{}.
\newblock \showarticletitle{Knowledge Enhanced Contextual Word
  Representations}. In \bibinfo{booktitle}{\emph{Proceedings of the 2019
  Conference on Empirical Methods in Natural Language Processing and the 9th
  International Joint Conference on Natural Language Processing
  (EMNLP-IJCNLP)}}. \bibinfo{pages}{43--54}.
\newblock


\bibitem[\protect\citeauthoryear{Qi, Yang, Liu, Dong, Sun, and Dong}{Qi
  et~al\mbox{.}}{2019}]%
        {qi2019openhownet}
\bibfield{author}{\bibinfo{person}{Fanchao Qi}, \bibinfo{person}{Chenghao
  Yang}, \bibinfo{person}{Zhiyuan Liu}, \bibinfo{person}{Qiang Dong},
  \bibinfo{person}{Maosong Sun}, {and} \bibinfo{person}{Zhendong Dong}.}
  \bibinfo{year}{2019}\natexlab{}.
\newblock \showarticletitle{OpenHowNet: An Open Sememe-based Lexical Knowledge
  Base}.
\newblock \bibinfo{journal}{\emph{arXiv preprint arXiv:1901.09957}}
  (\bibinfo{year}{2019}).
\newblock


\bibitem[\protect\citeauthoryear{Saxena, Tripathi, and Talukdar}{Saxena
  et~al\mbox{.}}{2020}]%
        {saxena2020improving}
\bibfield{author}{\bibinfo{person}{Apoorv Saxena}, \bibinfo{person}{Aditay
  Tripathi}, {and} \bibinfo{person}{Partha Talukdar}.}
  \bibinfo{year}{2020}\natexlab{}.
\newblock \showarticletitle{Improving multi-hop question answering over
  knowledge graphs using knowledge base embeddings}. In
  \bibinfo{booktitle}{\emph{Proceedings of the 58th annual meeting of the
  association for computational linguistics}}. \bibinfo{pages}{4498--4507}.
\newblock


\bibitem[\protect\citeauthoryear{Sun, Wang, Li, Feng, Chen, Zhang, Tian, Zhu,
  Tian, and Wu}{Sun et~al\mbox{.}}{2019}]%
        {sun2019ernie}
\bibfield{author}{\bibinfo{person}{Yu Sun}, \bibinfo{person}{Shuohuan Wang},
  \bibinfo{person}{Yukun Li}, \bibinfo{person}{Shikun Feng},
  \bibinfo{person}{Xuyi Chen}, \bibinfo{person}{Han Zhang},
  \bibinfo{person}{Xin Tian}, \bibinfo{person}{Danxiang Zhu},
  \bibinfo{person}{Hao Tian}, {and} \bibinfo{person}{Hua Wu}.}
  \bibinfo{year}{2019}\natexlab{}.
\newblock \showarticletitle{Ernie: Enhanced representation through knowledge
  integration}.
\newblock \bibinfo{journal}{\emph{arXiv preprint arXiv:1904.09223}}
  (\bibinfo{year}{2019}).
\newblock


\bibitem[\protect\citeauthoryear{Talmor and Berant}{Talmor and Berant}{2018}]%
        {talmor2018web}
\bibfield{author}{\bibinfo{person}{Alon Talmor} {and} \bibinfo{person}{Jonathan
  Berant}.} \bibinfo{year}{2018}\natexlab{}.
\newblock \showarticletitle{The Web as a Knowledge-Base for Answering Complex
  Questions}. In \bibinfo{booktitle}{\emph{Proceedings of the 2018 Conference
  of the North American Chapter of the Association for Computational
  Linguistics: Human Language Technologies, Volume 1 (Long Papers)}}.
  \bibinfo{pages}{641--651}.
\newblock


\bibitem[\protect\citeauthoryear{Wang, Tang, Duan, Wei, Huang, Ji, Cao, Jiang,
  and Zhou}{Wang et~al\mbox{.}}{2021b}]%
        {wang2021k}
\bibfield{author}{\bibinfo{person}{Ruize Wang}, \bibinfo{person}{Duyu Tang},
  \bibinfo{person}{Nan Duan}, \bibinfo{person}{Zhongyu Wei},
  \bibinfo{person}{Xuan-Jing Huang}, \bibinfo{person}{Jianshu Ji},
  \bibinfo{person}{Guihong Cao}, \bibinfo{person}{Daxin Jiang}, {and}
  \bibinfo{person}{Ming Zhou}.} \bibinfo{year}{2021}\natexlab{b}.
\newblock \showarticletitle{K-Adapter: Infusing Knowledge into Pre-Trained
  Models with Adapters}. In \bibinfo{booktitle}{\emph{Findings of the
  Association for Computational Linguistics: ACL-IJCNLP 2021}}.
  \bibinfo{pages}{1405--1418}.
\newblock


\bibitem[\protect\citeauthoryear{Wang, Gao, Zhu, Zhang, Liu, Li, and Tang}{Wang
  et~al\mbox{.}}{2021a}]%
        {wang2021kepler}
\bibfield{author}{\bibinfo{person}{Xiaozhi Wang}, \bibinfo{person}{Tianyu Gao},
  \bibinfo{person}{Zhaocheng Zhu}, \bibinfo{person}{Zhengyan Zhang},
  \bibinfo{person}{Zhiyuan Liu}, \bibinfo{person}{Juanzi Li}, {and}
  \bibinfo{person}{Jian Tang}.} \bibinfo{year}{2021}\natexlab{a}.
\newblock \showarticletitle{KEPLER: A unified model for knowledge embedding and
  pre-trained language representation}.
\newblock \bibinfo{journal}{\emph{Transactions of the Association for
  Computational Linguistics}}  \bibinfo{volume}{9} (\bibinfo{year}{2021}),
  \bibinfo{pages}{176--194}.
\newblock


\bibitem[\protect\citeauthoryear{Wang, Liu, and Shi}{Wang
  et~al\mbox{.}}{2017}]%
        {wang2017deep}
\bibfield{author}{\bibinfo{person}{Yan Wang}, \bibinfo{person}{Xiaojiang Liu},
  {and} \bibinfo{person}{Shuming Shi}.} \bibinfo{year}{2017}\natexlab{}.
\newblock \showarticletitle{Deep neural solver for math word problems}. In
  \bibinfo{booktitle}{\emph{Proceedings of the 2017 conference on empirical
  methods in natural language processing}}. \bibinfo{pages}{845--854}.
\newblock


\bibitem[\protect\citeauthoryear{Wei, Wang, Schuurmans, Bosma, Xia, Chi, Le,
  Zhou, et~al\mbox{.}}{Wei et~al\mbox{.}}{2022a}]%
        {wei2022chain}
\bibfield{author}{\bibinfo{person}{Jason Wei}, \bibinfo{person}{Xuezhi Wang},
  \bibinfo{person}{Dale Schuurmans}, \bibinfo{person}{Maarten Bosma},
  \bibinfo{person}{Fei Xia}, \bibinfo{person}{Ed Chi}, \bibinfo{person}{Quoc~V
  Le}, \bibinfo{person}{Denny Zhou}, {et~al\mbox{.}}}
  \bibinfo{year}{2022}\natexlab{a}.
\newblock \showarticletitle{Chain-of-thought prompting elicits reasoning in
  large language models}.
\newblock \bibinfo{journal}{\emph{Advances in Neural Information Processing
  Systems}}  \bibinfo{volume}{35} (\bibinfo{year}{2022}),
  \bibinfo{pages}{24824--24837}.
\newblock


\bibitem[\protect\citeauthoryear{Wei, Wang, Schuurmans, Bosma, Xia, Chi, Le,
  Zhou, et~al\mbox{.}}{Wei et~al\mbox{.}}{2022b}]%
        {weichain}
\bibfield{author}{\bibinfo{person}{Jason Wei}, \bibinfo{person}{Xuezhi Wang},
  \bibinfo{person}{Dale Schuurmans}, \bibinfo{person}{Maarten Bosma},
  \bibinfo{person}{Fei Xia}, \bibinfo{person}{Ed~H Chi},
  \bibinfo{person}{Quoc~V Le}, \bibinfo{person}{Denny Zhou}, {et~al\mbox{.}}}
  \bibinfo{year}{2022}\natexlab{b}.
\newblock \showarticletitle{Chain-of-Thought Prompting Elicits Reasoning in
  Large Language Models}. In \bibinfo{booktitle}{\emph{Advances in NeurIPS}}.
\newblock


\bibitem[\protect\citeauthoryear{Wu, Zhang, Fu, and Huang}{Wu
  et~al\mbox{.}}{2020}]%
        {wu2020knowledge}
\bibfield{author}{\bibinfo{person}{Qinzhuo Wu}, \bibinfo{person}{Qi Zhang},
  \bibinfo{person}{Jinlan Fu}, {and} \bibinfo{person}{Xuan-Jing Huang}.}
  \bibinfo{year}{2020}\natexlab{}.
\newblock \showarticletitle{A knowledge-aware sequence-to-tree network for math
  word problem solving}. In \bibinfo{booktitle}{\emph{Proceedings of the 2020
  conference on empirical methods in natural language processing (EMNLP)}}.
  \bibinfo{pages}{7137--7146}.
\newblock


\bibitem[\protect\citeauthoryear{Xiong, Du, Wang, and Stoyanov}{Xiong
  et~al\mbox{.}}{2020}]%
        {xiongpretrained}
\bibfield{author}{\bibinfo{person}{Wenhan Xiong}, \bibinfo{person}{Jingfei Du},
  \bibinfo{person}{William~Yang Wang}, {and} \bibinfo{person}{Veselin
  Stoyanov}.} \bibinfo{year}{2020}\natexlab{}.
\newblock \showarticletitle{Pretrained Encyclopedia: Weakly Supervised
  Knowledge-Pretrained Language Model}. In
  \bibinfo{booktitle}{\emph{International Conference on Learning
  Representations}}.
\newblock


\bibitem[\protect\citeauthoryear{Yao, Zhao, Yu, Du, Shafran, Narasimhan, and
  Cao}{Yao et~al\mbox{.}}{2022}]%
        {yao2022react}
\bibfield{author}{\bibinfo{person}{Shunyu Yao}, \bibinfo{person}{Jeffrey Zhao},
  \bibinfo{person}{Dian Yu}, \bibinfo{person}{Nan Du}, \bibinfo{person}{Izhak
  Shafran}, \bibinfo{person}{Karthik~R Narasimhan}, {and} \bibinfo{person}{Yuan
  Cao}.} \bibinfo{year}{2022}\natexlab{}.
\newblock \showarticletitle{ReAct: Synergizing Reasoning and Acting in Language
  Models}. In \bibinfo{booktitle}{\emph{The Eleventh International Conference
  on Learning Representations}}.
\newblock


\bibitem[\protect\citeauthoryear{Yasunaga, Bosselut, Ren, Zhang, Manning,
  Liang, and Leskovec}{Yasunaga et~al\mbox{.}}{2022}]%
        {yasunaga2022deep}
\bibfield{author}{\bibinfo{person}{Michihiro Yasunaga},
  \bibinfo{person}{Antoine Bosselut}, \bibinfo{person}{Hongyu Ren},
  \bibinfo{person}{Xikun Zhang}, \bibinfo{person}{Christopher~D Manning},
  \bibinfo{person}{Percy~S Liang}, {and} \bibinfo{person}{Jure Leskovec}.}
  \bibinfo{year}{2022}\natexlab{}.
\newblock \showarticletitle{Deep bidirectional language-knowledge graph
  pretraining}.
\newblock \bibinfo{journal}{\emph{Advances in Neural Information Processing
  Systems}}  \bibinfo{volume}{35} (\bibinfo{year}{2022}),
  \bibinfo{pages}{37309--37323}.
\newblock


\bibitem[\protect\citeauthoryear{Yasunaga, Ren, Bosselut, Liang, and
  Leskovec}{Yasunaga et~al\mbox{.}}{2021}]%
        {yasunaga2021qa}
\bibfield{author}{\bibinfo{person}{Michihiro Yasunaga}, \bibinfo{person}{Hongyu
  Ren}, \bibinfo{person}{Antoine Bosselut}, \bibinfo{person}{Percy Liang},
  {and} \bibinfo{person}{Jure Leskovec}.} \bibinfo{year}{2021}\natexlab{}.
\newblock \showarticletitle{QA-GNN: Reasoning with Language Models and
  Knowledge Graphs for Question Answering}. In
  \bibinfo{booktitle}{\emph{Proceedings of the 2021 Conference of the North
  American Chapter of the Association for Computational Linguistics: Human
  Language Technologies}}. \bibinfo{pages}{535--546}.
\newblock


\bibitem[\protect\citeauthoryear{Ye, Zhang, Deng, Chen, Chen, Xiong, Chen, and
  Chen}{Ye et~al\mbox{.}}{2022}]%
        {ye2022ontology}
\bibfield{author}{\bibinfo{person}{Hongbin Ye}, \bibinfo{person}{Ningyu Zhang},
  \bibinfo{person}{Shumin Deng}, \bibinfo{person}{Xiang Chen},
  \bibinfo{person}{Hui Chen}, \bibinfo{person}{Feiyu Xiong},
  \bibinfo{person}{Xi Chen}, {and} \bibinfo{person}{Huajun Chen}.}
  \bibinfo{year}{2022}\natexlab{}.
\newblock \showarticletitle{Ontology-enhanced Prompt-tuning for Few-shot
  Learning}. In \bibinfo{booktitle}{\emph{Proceedings of the ACM Web Conference
  2022}}. \bibinfo{pages}{778--787}.
\newblock


\bibitem[\protect\citeauthoryear{Yu, Zhu, Yang, and Zeng}{Yu
  et~al\mbox{.}}{2022}]%
        {yu2022jaket}
\bibfield{author}{\bibinfo{person}{Donghan Yu}, \bibinfo{person}{Chenguang
  Zhu}, \bibinfo{person}{Yiming Yang}, {and} \bibinfo{person}{Michael Zeng}.}
  \bibinfo{year}{2022}\natexlab{}.
\newblock \showarticletitle{Jaket: Joint pre-training of knowledge graph and
  language understanding}. In \bibinfo{booktitle}{\emph{Proceedings of the AAAI
  Conference on Artificial Intelligence}}, Vol.~\bibinfo{volume}{36}.
  \bibinfo{pages}{11630--11638}.
\newblock


\bibitem[\protect\citeauthoryear{Zhang, Zhu, Chen, Geng, Huang, Xu, Song, and
  Chen}{Zhang et~al\mbox{.}}{2023}]%
        {zhang2023structure}
\bibfield{author}{\bibinfo{person}{Wen Zhang}, \bibinfo{person}{Yushan Zhu},
  \bibinfo{person}{Mingyang Chen}, \bibinfo{person}{Yuxia Geng},
  \bibinfo{person}{Yufeng Huang}, \bibinfo{person}{Yajing Xu},
  \bibinfo{person}{Wenting Song}, {and} \bibinfo{person}{Huajun Chen}.}
  \bibinfo{year}{2023}\natexlab{}.
\newblock \showarticletitle{Structure Pretraining and Prompt Tuning for
  Knowledge Graph Transfer}. In \bibinfo{booktitle}{\emph{Proceedings of the
  ACM Web Conference 2023}}. \bibinfo{pages}{2581--2590}.
\newblock


\bibitem[\protect\citeauthoryear{Zhang, Bosselut, Yasunaga, Ren, Liang,
  Manning, and Leskovec}{Zhang et~al\mbox{.}}{2022}]%
        {zhang2022greaselm}
\bibfield{author}{\bibinfo{person}{Xikun Zhang}, \bibinfo{person}{Antoine
  Bosselut}, \bibinfo{person}{Michihiro Yasunaga}, \bibinfo{person}{Hongyu
  Ren}, \bibinfo{person}{Percy Liang}, \bibinfo{person}{Christopher~D Manning},
  {and} \bibinfo{person}{Jure Leskovec}.} \bibinfo{year}{2022}\natexlab{}.
\newblock \showarticletitle{GreaseLM: Graph REASoning Enhanced Language
  Models}. In \bibinfo{booktitle}{\emph{International conference on learning
  representations}}.
\newblock


\bibitem[\protect\citeauthoryear{Zhang, Han, Liu, Jiang, Sun, and Liu}{Zhang
  et~al\mbox{.}}{2019}]%
        {zhang2019ernie}
\bibfield{author}{\bibinfo{person}{Zhengyan Zhang}, \bibinfo{person}{Xu Han},
  \bibinfo{person}{Zhiyuan Liu}, \bibinfo{person}{Xin Jiang},
  \bibinfo{person}{Maosong Sun}, {and} \bibinfo{person}{Qun Liu}.}
  \bibinfo{year}{2019}\natexlab{}.
\newblock \showarticletitle{ERNIE: Enhanced Language Representation with
  Informative Entities}. In \bibinfo{booktitle}{\emph{Proceedings of the 57th
  Annual Meeting of the Association for Computational Linguistics}}.
  \bibinfo{pages}{1441--1451}.
\newblock


\bibitem[\protect\citeauthoryear{Zhao, Zhou, Gong, Zhang, Zhou, Sha, Chen,
  Wang, Liu, and Wen}{Zhao et~al\mbox{.}}{2022}]%
        {zhao2022jiuzhang}
\bibfield{author}{\bibinfo{person}{Wayne~Xin Zhao}, \bibinfo{person}{Kun Zhou},
  \bibinfo{person}{Zheng Gong}, \bibinfo{person}{Beichen Zhang},
  \bibinfo{person}{Yuanhang Zhou}, \bibinfo{person}{Jing Sha},
  \bibinfo{person}{Zhigang Chen}, \bibinfo{person}{Shijin Wang},
  \bibinfo{person}{Cong Liu}, {and} \bibinfo{person}{Ji-Rong Wen}.}
  \bibinfo{year}{2022}\natexlab{}.
\newblock \showarticletitle{JiuZhang: A Chinese Pre-trained Language Model for
  Mathematical Problem Understanding}. In \bibinfo{booktitle}{\emph{Proceedings
  of the 28th ACM SIGKDD Conference on Knowledge Discovery and Data Mining}}.
  \bibinfo{pages}{4571--4581}.
\newblock


\bibitem[\protect\citeauthoryear{Zhou, Sch{\"a}rli, Hou, Wei, Scales, Wang,
  Schuurmans, Cui, Bousquet, Le, et~al\mbox{.}}{Zhou et~al\mbox{.}}{2022}]%
        {zhou2022least}
\bibfield{author}{\bibinfo{person}{Denny Zhou}, \bibinfo{person}{Nathanael
  Sch{\"a}rli}, \bibinfo{person}{Le Hou}, \bibinfo{person}{Jason Wei},
  \bibinfo{person}{Nathan Scales}, \bibinfo{person}{Xuezhi Wang},
  \bibinfo{person}{Dale Schuurmans}, \bibinfo{person}{Claire Cui},
  \bibinfo{person}{Olivier Bousquet}, \bibinfo{person}{Quoc~V Le},
  {et~al\mbox{.}}} \bibinfo{year}{2022}\natexlab{}.
\newblock \showarticletitle{Least-to-Most Prompting Enables Complex Reasoning
  in Large Language Models}. In \bibinfo{booktitle}{\emph{The Eleventh
  International Conference on Learning Representations}}.
\newblock


\bibitem[\protect\citeauthoryear{Zhu, Xu, Ren, Lin, Jiang, and Yu}{Zhu
  et~al\mbox{.}}{2023}]%
        {zhu2023knowledge}
\bibfield{author}{\bibinfo{person}{Chenguang Zhu}, \bibinfo{person}{Yichong
  Xu}, \bibinfo{person}{Xiang Ren}, \bibinfo{person}{Bill~Yuchen Lin},
  \bibinfo{person}{Meng Jiang}, {and} \bibinfo{person}{Wenhao Yu}.}
  \bibinfo{year}{2023}\natexlab{}.
\newblock \showarticletitle{Knowledge-augmented methods for natural language
  processing}. In \bibinfo{booktitle}{\emph{Proceedings of the Sixteenth ACM
  International Conference on Web Search and Data Mining}}.
  \bibinfo{pages}{1228--1231}.
\newblock


\end{thebibliography}

\appendix

\begin{table*}
  \caption{Statistics of Datasets}
  \label{tab:dataset_statistics}
  \centering
  \begin{tabular}{c|cccc}
    \toprule
    Dataset & CN-QA & ComplexWebQuestions & FreebaseQA & Math23K \\
    KG & CN-KG & Wikidata & Wikidata & HowNet \\
    \midrule
    \#Questions & 13,041 & 13,544 & 15,811 & 23,162 \\
    \#Simple questions & 12,265 & 0 & 13,070 & / \\
    \#Hard questions & 776 & 13,544 & 2,741 & / \\
    Avg. answer per question & 1.67 & 1.43 & 1 & 1 \\
    \midrule
    \#Entity & 1,477,923 & 397,133 & 397,133 & 237,861 \\
    \#Relations \& attributes & 1,112 & 733 & 733 & 6 \\
    \#All triples & 6,352,980 & 2,900,156 & 2,900,156 & 1,206,695 \\
    \#Relation triples & 4,081,756 & 2,900,156 & 2,900,156 & 1,206,695 \\
    \#Attribute triples & 2,271,224 & 0 & 0 & 0 \\
    \midrule
    \# Corpus for lesson 1 & 6,352,980 & 2,900,156 & 2,900,156 & 1,206,695 \\
    \# Corpus for lesson 2 & 1,806,861 & 3,128,153 & 3,128,153 & 1,356,960 \\
    \# Corpus for lesson 3 & 1,806,861 & 3,128,153 & 3,128,153 & 1,356,960 \\
    \bottomrule
  \end{tabular}
\end{table*}

% \section{Architecture of Knowledge Adapter}\label{app:implementation}

% \begin{figure}[t]
%   \centering
%   \includegraphics[width=180px]{fig/KA}
%   \caption{An implementation of the knowledge adaptation module (KA) using BERT with additional inputs.}
%   \Description{An implementation of the knowledge adaptation module (KA) using BERT with additional inputs.}
%   \label{fig:ka}
% \end{figure}

% The architecture of the KA module implemented in section~\ref{sec:implementation} is demonstrated in Figure~\ref{fig:ka}, where $s_i$ represents the semantic vector from the $i$th layer of the original language model $\mathit{LM}$, and $h_i$ represents the hidden knowledge-enhanced vector output of the $i$th layer of the knowledge adapter $\mathit{Ad}$. $\mathit{AL}$ means the adapter layer, $\mathit{Trm}^L$ and $\mathit{Trm}^K$ mean transformer blocks with different hidden dimension, and $\mathit{Ln}^I$ and $\mathit{Ln}^O$ mean linear models.

\section{Datasets}\label{app:dataset}

\textbf{CN-QA} is a Chinese KBQA dataset collected from smart voice assistant accompanied by a KG named CN-KG with both entity relations and attributes. \textbf{ComplexWebQuestions}~\cite{talmor2018web} is a public KBQA dataset with complex questions built on WebQuestions and Freebase. \textbf{FreebaseQA}~\cite{jiang2019freebaseqa} is another public KBQA dataset based on Freebase with both simple and complex questions derived from TriviaQA and trivia websites. Since Freebase has been merged to Wikidata, we use the Wikidata dump in~\cite{wang2021kepler}, and map entities to Wikidata to construct an answerable subset for ComplexWebQuestions and FreebaseQA. \textbf{Math23K}~\cite{wang2017deep} is a public generative math word problem dataset which answers the question with a generated mathematical expression. We construct a KG based on the semantic web HowNet~\cite{qi2019openhownet} for Math23K following~\cite{wu2020knowledge}. 
The statistics of the datasets are available in Table~\ref{tab:dataset_statistics}. 

\section{Introduction to Baselines}\label{app:baseline}
The introduction to the baselines are listed as follows.
\begin{itemize}
  \item \textbf{BERT}~\cite{devlin2019bert} was the most widely used pretrained language model, based on which our framework is implemented, thus we add BERT as baseline to evaluate the improvement.
  \item \textbf{RoBERTa}~\cite{liu2019roberta} studied the impacts of hyperparameters and task design in pretraining, and achieved a robustly optimized BERT with significant improvements.
  \item \textbf{ERNIE}~\cite{zhang2019ernie} developed an aggregator network to explicitly combine the entity embedding learned from KG with the semantics learned by LM to inject knowledge into the LM.
  \item \textbf{K-BERT}~\cite{liu2020k} directly linked the related KG triples with the sentence to inject the knowledge, which was fed to the LM together for the knowledge-enhanced representation.
  \item \textbf{KEPLER}~\cite{wang2021kepler} trained the LM as the knowledge embedding model, where the entity embedding was generated by the LM on the entity description.
  \item \textbf{K-Adapter}~\cite{wang2021k} designed a neural adapter for each kind of infused knowledge, and trained the adapters on different knowledge pretraining tasks.
  \item \textbf{EmbedKGQA}~\cite{saxena2020improving} represented the question and KG in the same latent space, and inferred the answer with simple vector computation. 
  \item \textbf{GPT4}~\cite{achiam2023gpt} is the state-of-the-art LLMs developed by OpenAI, which provides API to access the service.
  \item \textbf{ChatGLM2}~\cite{du2022glm} is an open-sourced bilingual LLMs with good performance, with its 6B pretrained weights released.
\end{itemize}

\section{More Cases}\label{app:cases}

\begin{table*}
  \caption{More Cases Predicted by KICP and BERT}
  \label{tab:more_cases}
  \centering
  \begin{tabular}{c|l}
    \toprule
    Category & Cases \\
    \midrule
    \multirow{12}{*}{Easy} & \textbf{Case 1:} Aberystwyth lies on which bay ?\\
    & \textbf{KICP:} Cardigan (\textbf{correct}) \\
    & \textbf{BERT:} Blaenau Gwent (\textbf{wrong}) \\
    \cmidrule(lr){2-2}
    & \textbf{Case 2:} In \textit{Alice in Wonderland}, who wanted to decapitate anyone who offended her ?\\
    & \textbf{KICP:} Queen of Hearts (\textbf{correct}) \\
    & \textbf{BERT:} Daisy Fay (\textbf{wrong}) \\
    \cmidrule(lr){2-2}
    & \textbf{Case 3:} Who wrote the thriller novel \textit{Birds of Prey} ?\\
    & \textbf{KICP:} Wilbur Smith (\textbf{correct}) \\
    & \textbf{BERT:} Ludwig von Mises (\textbf{wrong}) \\
    \cmidrule(lr){2-2}
    & \textbf{Case 4:} Io, Europa, Ganymede and Callisto are all moons of which planet in our solar  system ?\\
    & \textbf{KICP:} Jupiter (\textbf{correct}) \\
    & \textbf{BERT:} Pluto (\textbf{wrong}) \\
    \midrule
    \multirow{9}{*}{Hard} & \textbf{Case 5:} What kind of money does the country with the nation anthem \textit{Du gamla, Du fria}  use ?\\
    &\textbf{KICP:} Swedish Krona (\textbf{correct}) \\
    &\textbf{BERT:} / (\textbf{wrong}) \\
    \cmidrule(lr){2-2}
    &\textbf{Case 6:} What form of government is used in the country that uses Chilean Peso ?\\
    &\textbf{KICP:} Presidential system | Unitary state (\textbf{correct}) \\
    &\textbf{BERT:} Presidential system | Unitary state | Patrimonial monarchy (\textbf{wrong}) \\
    \cmidrule(lr){2-2}
    &\textbf{Case 7:} What is the nationality of the author of \textit{The Little Prince} ?\\
    &\textbf{KICP:} France (\textbf{correct}) \\
    &\textbf{BERT:} America (\textbf{wrong}) \\
    \midrule
    \multirow{8}{*}{Wrong} & \textbf{Case 8:} Which comedy actor played Charlie Bind in the 1964 film \textit{Carry on Spying} ?\\
    &\textbf{KICP:} Peter Hinwood (\textbf{wrong}) \\
    &\textbf{BERT:} Peter Hinwood (\textbf{wrong}) \\
    &\textbf{Answer:} Charles Hawtrey \\
    \cmidrule(lr){2-2}
    &\textbf{Case 9:} What team did Drogba play for that won the 2014 Coupe de France Final championship ?\\
    &\textbf{KICP:} Piast Gliwice (\textbf{wrong}) \\
    &\textbf{BERT:} Germinal Beerschot (\textbf{wrong}) \\
    &\textbf{Answer:} En Avant de Guingamp \\
    \bottomrule
  \end{tabular}
\end{table*}

We also provide more cases predicted by KICP and BERT on the KBQA datasets in Table~\ref{tab:more_cases} in addition to section~\ref{sec:case_study}. We classify these cases into three categories, i.e., the easy questions, hard questions, and wrong questions that both KICP and BERT fail. We can summarize the following observations. First, the easy questions can be answered with only one knowledge triples, which investigates whether the LM can memorize and exploit the knowledge. From the cases, KICP performs better than BERT. Next, the hard questions require reasoning over multiple knowledge facts. There are two typical mistakes in these cases, i.e., wrong answers (case 6 and 7) and failed prediction (case 5), which shows that the method may be not so capable of effective reasoning. Last, there are also questions mistakenly answered by KICP (case 8 and 9). In these cases, both the two methods make similar wrong prediction, which shows that there are still much room to improve for KICP, such as more reasoning patterns and more efficient knowledge learning and exploitation. 

\section{Samples of Corpus}\label{app:corpus}

\begin{table*}
  \caption{Samples of the Constructed Corpus in the CR Module}
  \label{tab:corpus_samples}
  \centering
  \begin{tabular}{c|cl}
    \toprule
    Lesson && Samples \\
    \midrule
    \multirow{10}{*}{Lesson 1} &(1)& [CLS] sir frederick ashton nationality united kindom [SEP] \\
    && [CLS] [MASK] [MASK] [MASK] nationality united kindom [SEP] \\
    \cmidrule(lr){2-3}
    &(2)& [CLS] wilhelm friedrich kuhne member of royal society [SEP] \\
    && [CLS] wilhelm friedrich kuhne member of [MASK] [MASK] [SEP] \\
    \cmidrule(lr){2-3}
    &(3)& [CLS] republic of maldives used money maldivian rufiyah [SEP] \\
    && [CLS] republic of maldives [MASK] [MASK] maldivian rufiyah [SEP] \\
    \cmidrule(lr){2-3}
    &(4)& [CLS] sarbogard district time euro time [SEP] \\
    && [CLS] sarbogard district time [MASK] orthogonal [SEP] \\
    \cmidrule(lr){2-3}
    &(5)& [CLS] first hellenic republic flag flag of greece [SEP] \\
    && [CLS] [MASK] [MASK] [MASK] flag flag of greece [SEP] \\
    \midrule
    \multirow{17}{*}{Lesson 2} &(6)& [CLS] collaroy plateau based in p : nsw [SEP] au - ns divides into gundagai  shire council [SEP] collaroy plateau \\ && based in divides into gundagai shire  council [SEP] \\
    &&\\
    && [CLS] collaroy plateau based in p : nsw [SEP] au - ns [MASK] into gundagai  shire  council [SEP] collaroy plateau \\ && based in [MASK] [MASK] gundagai shire  council [SEP] \\
    \cmidrule(lr){2-3}
    &(7)& [CLS] star fox 64 3d part of the series star fox ( virtual boy ) [SEP] starfox ( virtual boy ) characters fox makuraudo \\ && [SEP] fox mccloud recording by ohara  takashi [SEP] star fox 64 3d part of the series  characters recording by ohara \\ &&  takashi [SEP] \\
    &&\\
    && [CLS] star fox 64 3d part of the series [MASK] fox [MASK] [MASK] [MASK] ) [SEP] starfox ( virtual boy ) [MASK] \\ && fox makuraudo [SEP] [MASK]  [MASK] [MASK] [MASK] recording by ohara takashi [SEP] star fox  64 3d part of the \\ && series [MASK] recording by ohara takashi [SEP] \\
    \cmidrule(lr){2-3}
    &(8)& [CLS] spannarhyttan timezone utc + 2 : 00 [SEP] spannarhyttan timezone utc + 1 : 00  [SEP] spannarhyttan timezone \\ && utc + 2 : 00 utc + 1 : 00 [SEP]\\
    &&\\
    && [CLS] spannarhyttan timezone utc [MASK] [MASK] : [MASK] [SEP] spannarhyttan  [MASK] [MASK] utc + 1 : 00 \\ && [SEP] spannarhyttan \#\#unes \#\#zone  [MASK] [MASK]  133 : [MASK] utc + 1 : 00 [SEP]\\
    \midrule
    \multirow{8}{*}{Lesson 3} &(9)& [CLS] theobald ziegler working at on lake the rhine [SEP] \\
    % && \\
    && [CLS] theobald ziegler working at on lake [MASK] [MASK] [SEP] \\
    % && \\
    && ( [CLS] theobald ziegler working at strassbourg [SEP] \\
    && \ \ [CLS] strassbourg on lake the rhine [SEP] )\\
    \cmidrule(lr){2-3}
    &(10)& [CLS] ferrieres , somme shares border with ailly - sur - somme pont - de - metz [SEP] \\
    % && \\
    && [CLS] ferrieres , somme [MASK] [MASK] [MASK] ailly - sur - somme pont - de - metz [SEP] \\
    % &&\\
    && ( [CLS] ferrieres , somme shares border with ailly - sur - somme [SEP] \\
    && \ \ [CLS] ferrieres , somme shares border with pont - de - metz [SEP] ) \\
    \bottomrule
  \end{tabular}
\end{table*}

We demonstrate some samples of the constructed corpora for the three lessons of the CR module in Table~\ref{tab:corpus_samples}. We place the unmasked version of each sentence on first line and masked one on second, and recover the split words for readability. The sentences are all in lower cases due to tokenization. We also provide related knowledge in the last two lines for lesson 3 for readability as some key information may be discarded. 

\end{document}